\documentclass[10pt,twocolumn,letterpaper]{article}

\usepackage{cvpr}
\usepackage{times}
\usepackage{epsfig}
\usepackage{graphicx}
\usepackage{amsmath}
\usepackage{amssymb}

\usepackage[breaklinks=true,bookmarks=false]{hyperref}
\graphicspath{{Figs1/}}
\usepackage{caption}
\newcommand{\x}{\times}
\cvprfinalcopy 

\def\cvprPaperID{3404} 

\begin{document}
\title{Uncertainty Guided Multi-Scale Residual Learning-using a Cycle Spinning CNN for Single Image De-Raining}

\author{Rajeev Yasarla and Vishal M. Patel\\
	Johns Hopkins University\\
	Department of Electrical and Computer Engineering, Baltimore, MD 21218, USA\\
	{\tt\small ryasarl1@jhu.edu, vpatel36@jhu.edu}
}

\maketitle

\begin{abstract}
	
	Single image de-raining is an extremely challenging problem since the rainy image may contain rain streaks which may vary in size, direction and density. Previous approaches have attempted to address this problem by leveraging some prior information to remove rain streaks from a single image.   One of the major limitations of these approaches is that they do not consider the location information of rain drops in the image.  The proposed Uncertainty guided Multi-scale Residual Learning (UMRL) network attempts to  address this issue by learning the rain content at different scales and using them to estimate the final de-rained output.  In addition, we introduce a technique which guides the network to learn the network weights based on the confidence measure about the estimate.  Furthermore, we introduce a new training and testing procedure based on the notion of cycle spinning to improve the final de-raining performance.  Extensive experiments on synthetic and real datasets to demonstrate that the proposed method achieves significant improvements over the recent state-of-the-art methods. Code is available at: \url{https://github.com/rajeevyasarla/UMRL--using-Cycle-Spinning}
	
\end{abstract}


\section{Introduction}
Many practical computer vision-based systems such as surveillance and autonomous driving often require processing and analysis of videos and images captured under adverse weather conditions such as rain, snow, haze etc.  These weather-based conditions adversely affect the visual quality of images and as a result often degrade the performance of vision systems. Hence, it is important to develop algorithms that can automatically remove these artifacts before they are fed to a vision-based system for further processing. 

\begin{figure}[t!]
	
	\centering
	\includegraphics[width=1.305cm,height=0.98cm]{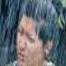}
	\includegraphics[width=1.305cm,height=0.98cm]{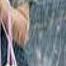}
	\includegraphics[width=1.305cm,height=0.98cm]{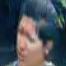}
	\includegraphics[width=1.305cm,height=0.98cm]{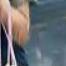}
	\includegraphics[width=1.305cm,height=0.98cm]{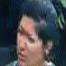}
	\includegraphics[width=1.305cm,height=0.98cm]{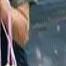}
	\\
	\includegraphics[width=2.7cm,height=1.58cm]{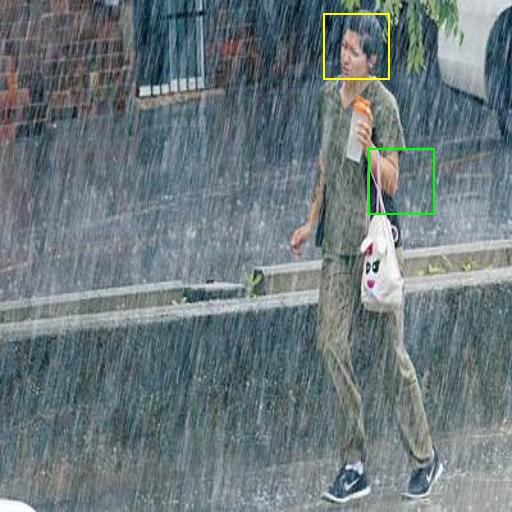}
	\includegraphics[width=2.7cm,height=1.58cm]{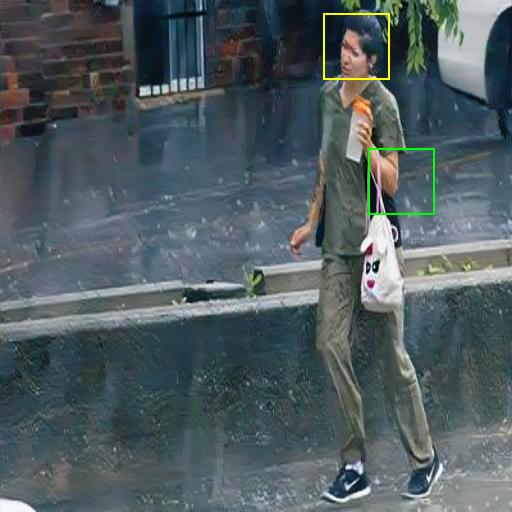}
	\includegraphics[width=2.7cm,height=1.58cm]{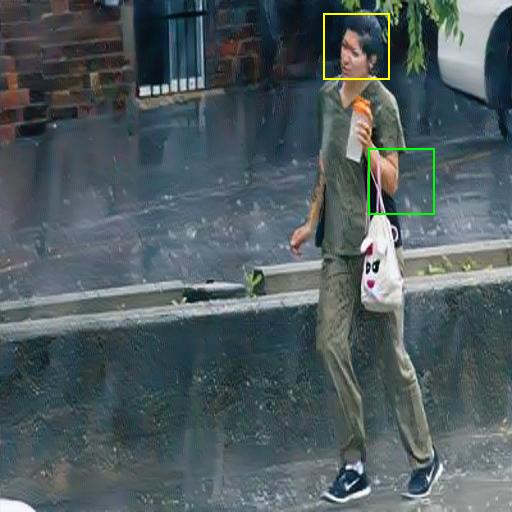}\\
	(a)\hskip70pt(b)\hskip70pt(c)\\
	\centering
	\includegraphics[width=1.305cm,height=0.98cm]{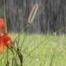}
	\includegraphics[width=1.305cm,height=0.98cm]{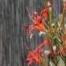}
	\includegraphics[width=1.305cm,height=0.98cm]{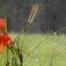}
	\includegraphics[width=1.305cm,height=0.98cm]{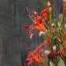}
	\includegraphics[width=1.305cm,height=0.98cm]{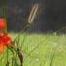}
	\includegraphics[width=1.305cm,height=0.98cm]{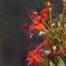}
	\\
	\includegraphics[width=2.7cm,height=1.58cm]{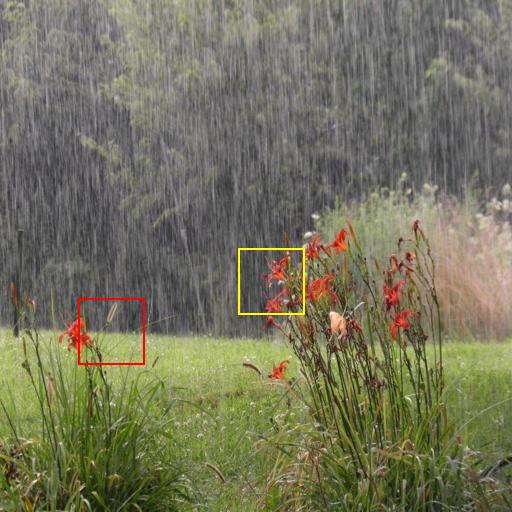}
	\includegraphics[width=2.7cm,height=1.58cm]{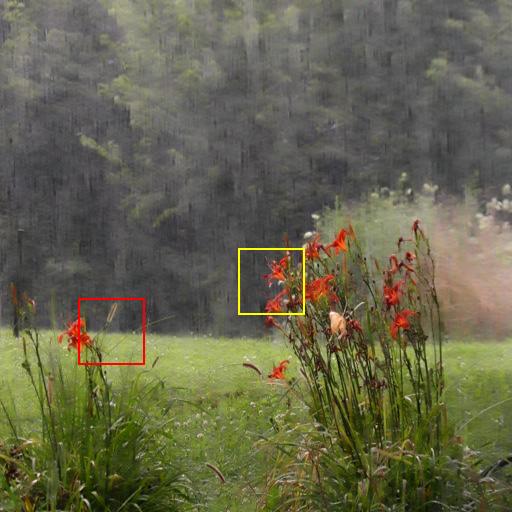}
	\includegraphics[width=2.7cm,height=1.58cm]{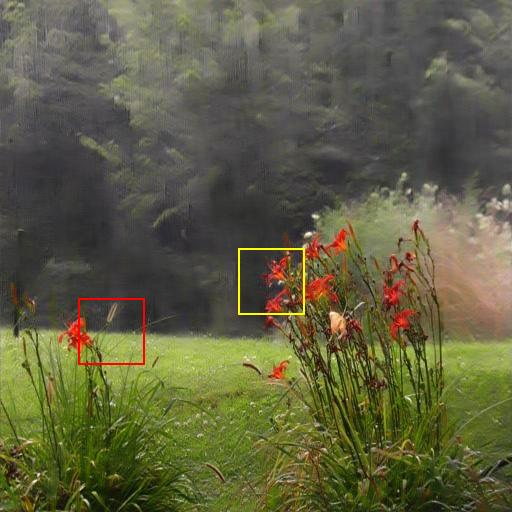}\\
	(d)\hskip70pt(e)\hskip70pt(f)
	\vskip-10pt
	\caption{Sample image de-raining results.  (a) Rainy image. (b)  De-rained using DID-MDN \cite{Authors18} where zoomed in part shows the blurry effects on face and various rain streaks near the elbow.  (c) De-rained using UMRL. (d) Rainy image. (e) De-rained using Fu et al. \cite{Authors17d} where zoomed in part shows under de-raining of the image. (f) De-rained using UMRL, zoomed in highlighted parts show the clear differences between UMRL and other compared methods.}
	\label{Fig:Initial}
	\vskip-20pt
\end{figure}

In this paper, we address the problem of removing rain streaks from a single rainy image.  Rain streak removal or image de-raining is a difficult problem since a rainy image may contain rain streaks which may vary in size, direction and density.  A number of different techniques have been developed in the literature to address this problem.   These algorithms can be clustered into two main groups - (i) video based algorithms \cite{Authors6,Authors7,Authors15c,Authors18e,Authors18f}, and (ii) single image-based algorithms \cite{Authors18,Authors17f,Authors16, Authors17h, Authors17c}. Algorithms corresponding to the first category assume temporal consistency among the image frames, and use this assumption for de-raining. On the other hand, single image de-raining methods attempt to use some prior information to remove rain components from a single image \cite{Authors16, Authors12,Authors17c,Authors18}.  Priors such as sparsity \cite{Authors17g, Authors15} and low-rank representation \cite{Low_Rank_ICCV2013} have been used in the literature.  In particular, the method proposed by Fu et al. \cite{Authors17d} uses a priori image domain knowledge by focusing on high frequency details during training to improve the de-raining performance.  However, it was shown in \cite{Authors18}, that this method tends to remove some important parts  in  the  de-rained  image (see Figure~\ref{Fig:Initial}(e)).   Similarly, a recent work by Zhang  and Patel \cite{Authors18} uses the image-level priors to estimate the rain density information which is then used for de-raining.  Although their approach provides the state-of-the-art results, they estimate image level priors which do not consider the location information of rain drops in the image.  As a result, their algorithm tends to introduce some artifacts in the final de-rained images.  These artifacts can be clearly seen from the de-rained results shown in Figure~\ref{Fig:Initial}(b).  

In this paper we take a different approach to image de-raining where we make use of the observation that rain streak density and direction does not change drastically with different scales. Rather than relying on the rain density information (i.e. heavy, medium or light) present in the rainy image \cite{Authors18}, we develop a method in which the rain streak location information is taken in to consideration in a multi-scale fashion  to improve the de-raining performance.   While providing the estimated rain content (i.e. residual map) to the subsequent layers of the network, we may end-up propagating the errors in estimations. To block the flow of incorrect estimation in  rain streaks, we estimate an uncertainty metric along with the rain streak information. We use an Unet architecture with skip connections \cite{Authors15b} as our base network. The proposed network learns the residue at each level in the decoder of Unet with an uncertainty map, which indicates how confident the network is about the rain content it learned. Say there are $L$ layers in the decoder network, the uncertainty map generated at layer  $\:``l"$ is given to layer $\:``l+1"$ so that the subsequent layers of $\:``l"$ can discard the rain content learned by layer $\:``l"$ if the confidence value is low in the uncertainty map.

Another important contribution of our work is that we propose to incorporate the cycle spinning framework of Coifman and Donoho \cite{Authors95} into our de-raining method.  Cycle spinning was originally proposed to remove the artifacts introduced by orthogonal wavelets in image de-noising.  Similar to wavelets, deep learning-based methods also introduce some artifacts near the edges  of the de-rained images (see Figure~\ref{Fig:Initial}).  In cycle spinning, the data is first shifted by some amount,  the shifted data are then de-noised, the de-noised data are then un-shifted, and finally the un-shifted data are averaged to obtain the final de-noised result.  Cycle spinning has been successfully applied to reduce the artifacts introduced near the edges in many applications including image de-blurring \cite{Dono2005} and de-noising \cite{Authors95}, \cite{Authors14_1}.  Hence, we adopt it in our de-raining framework. In fact, we show that cycle spinning is a generic method that can be used to improve the performance of any deep learning-based image de-raining method.  

Figure~\ref{Fig:Initial} (c) and (f) present sample results from our Uncertainty guided Multi-scale Residual Learning using cycle spinning (UMRL) network, where one can clearly see that UMRL is able to remove the noise artifacts and provides better results as compared to \cite{Authors18} and \cite{Authors17f}.

To summarize, this paper makes the following  contributions:
\begin{itemize}
	\item A novel method called UMRL is proposed which generates the rain streak content at each location of the image along with the uncertainty map that guides the subsequent layers about the rain streak information at each location.
	\item We incorporate cycle spinning in both training and testing phases of our network to improve the final de-raining performance.
	\item We run extensive experiments to show the performance of UMRL against the several recent state-of the-art approaches on both synthetic and real rainy images. Furthermore, an ablation study is conducted to demonstrate the effectiveness of different parts of the proposed UMRL network.
\end{itemize}

\begin{figure*}[htp!]
	\begin{center}
		\centering
		\includegraphics[width=12cm,height=6.3cm]{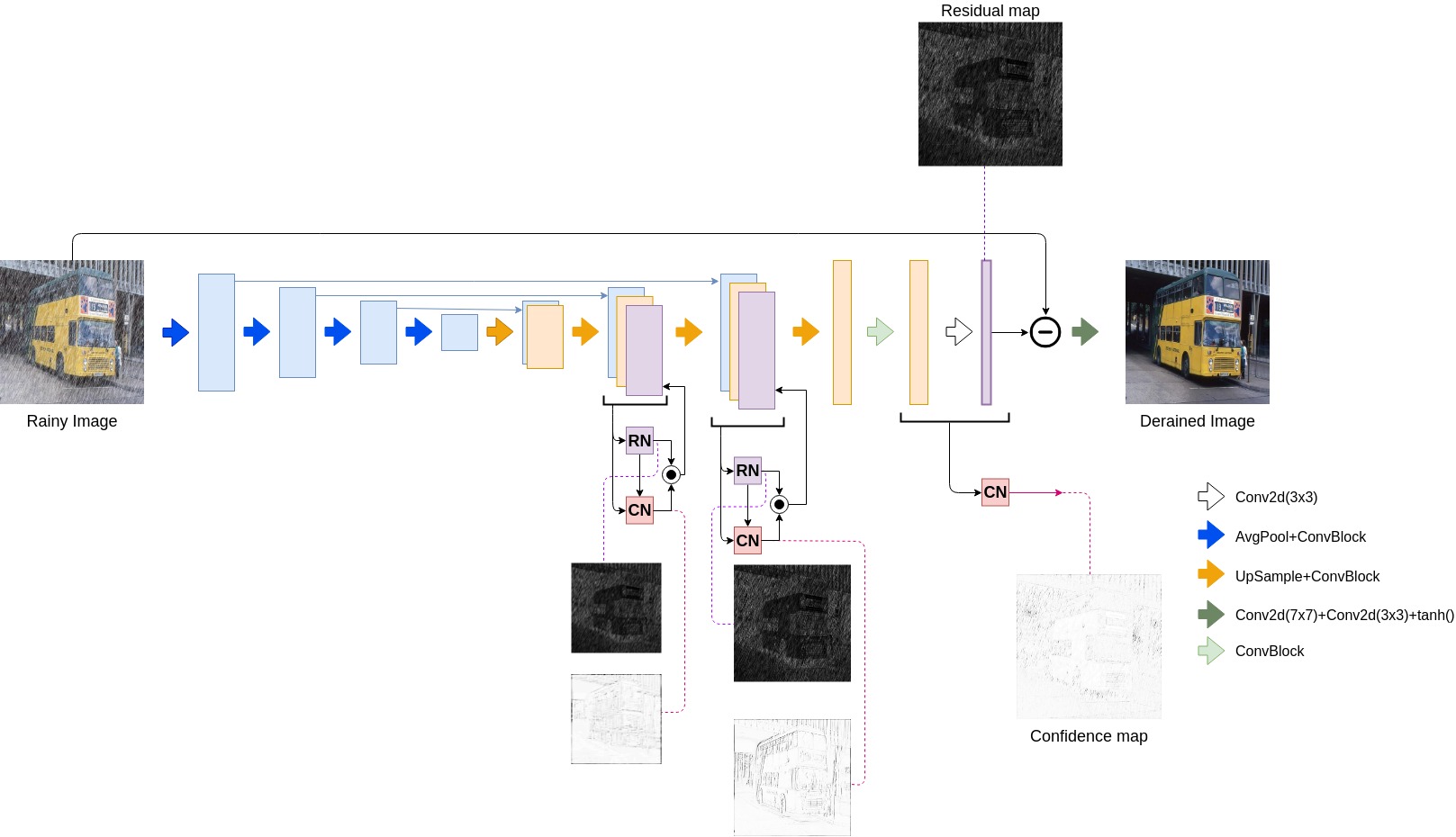}\\
		\caption{An overview of the proposed UMRL network.  The aim of the UMRL network is to estimate the clean image given the corresponding rainy image. To address that, UMRL learns the residual maps and computes the confidence maps to guide the network. To achieve this, we introduce RN and CN networks and feed their outputs to the subsequent layers.}
		\label{Fig:UMRL}
	\end{center}
	\vskip-25pt
\end{figure*}

\section{Background and Related Work} 
An observed rainy image $y$ can be modeled as the superposition of a rain component (i.e. residual map) $r$ with a clean image $x$ as follows
\setlength{\belowdisplayskip}{0pt} \setlength{\belowdisplayshortskip}{0pt}
\setlength{\abovedisplayskip}{0pt} \setlength{\abovedisplayshortskip}{0pt}
\begin{equation}
y = x + r.
\label{eq:formulation}
\end{equation}
Given $ y $ the goal of image de-raining is to estimate $x$.  This can be done by first estimating the residual map $r$ and then subtracting it from the observed image $y$.  Various methods have been proposed in the literature for image de-raining \cite{Authors12,Authors14d,Authors15,Authors14b,Authors16} including dictionary learning-based \cite{Authors9}, Gaussian mixture-model (GMM) based \cite{Authors2000}, and low-rank representation based \cite{Authors13} methods.   In recent years, deep learning-based single image de-raining methods have also been proposed in the literature.  Fu et al. \cite{Authors17d} proposed a convolutional neural network (CNN) based approach in which they directly learn the  mapping  relationship  between  rainy  and  clean  image  detail layers  from  data. Zhang et al. \cite{Authors17e} proposed a generative adversarial network (GAN) based method for image de-raining.  Furthermore, to  minimize  the artifacts  introduced  by  GANs and  ensure  better  visual  quality,  a  new  refined  loss  function was also  introduced in \cite{Authors17e}.  Fu et al. \cite{Authors17f}  presented an end-to-end deep learning framework for removing rain from individual images using a deep detail network which directly reduces the mapping range from input to output.  Zhang and Patel \cite{Authors18}  proposed  a density-aware multi-stream densely connected CNN for joint rain density  estimation  and  de-raining.  Their  network  automatically  determines  the rain-density  information and  then  efficiently  removes  the corresponding  rain-streaks  using the estimated  rain-density label.  Note the methods proposed in \cite{Authors17f}, and \cite{Authors18} showed the benefits of using multi-scale networks for image de-raining.   Recently, Wang et al. \cite{Authors17b} proposed a hierarchical approach based on estimating different frequency details of an image to get the de-rained image.   The method proposed by Qian et al. \cite{Authors18b} generates attentive maps using the recurrent neural networks, and then uses the features from different scales to compute the loss for removing the rain drops on glasses.   Note that this method was specifically designed for removing rain drops from a glass rather than removing rain streaks from an image. \cite{NonLocal2018,Xu_2018_CVPR,Authors18d} illustrated the importance of attention based methods in low-level vision tasks. In a recent work \cite{Authors18d}, Li et al. proposed a convolutional and recurrent neural network-based method for single image de-raining which makes use of the contextual information for rain removal.   It was observed in \cite{Authors18}, that some of the recent deep learning-based methods tend to under de-rain or over de-rain  the image if the rain condition present in the rainy image is not properly considered during training.  

\section{Proposed Method}
Unlike many deep learning-based methods that directly estimate the de-rained image from the noisy observation, we take a different approach in which we first estimate the rain streak component $\hat{r}$ (i.e. residual map) and then use it to estimate the de-rained image as  $\hat{x} = y - \hat{r}$. We define $ c $ as the confidence score which is an uncertainty map about the estimation of $\hat{r}$.  The confidence score at each pixel is a measure of how much the network is certain about the residual value computed at each pixel. Qian et al. \cite{Authors18b} estimate an attentive map based on the rainy image using a recurrent network and then use it as a location-based information to the de-raining network. In contrast, our method combines the residual and confidence information judiciously and uses them as input to subsequent layers at higher scales.  In this way, it passes the location-based rain information to the rest of the network. We estimate the residual map and its corresponding uncertainty map at three different scales, $\:\{\hat{r}_{\x1},c_{\x1}\} $ (at the original input size), $\:\{\hat{r}_{\x2},c_{\x2}\} $ (at 0.5 scale of input size), and $\:\{\hat{r}_{\x4},c_{\x4}\} $ (at 0.25 scale of input size). 

\begin{figure}[htp!]
	\centering
	\includegraphics[width=2.7cm,height = 1.58cm]{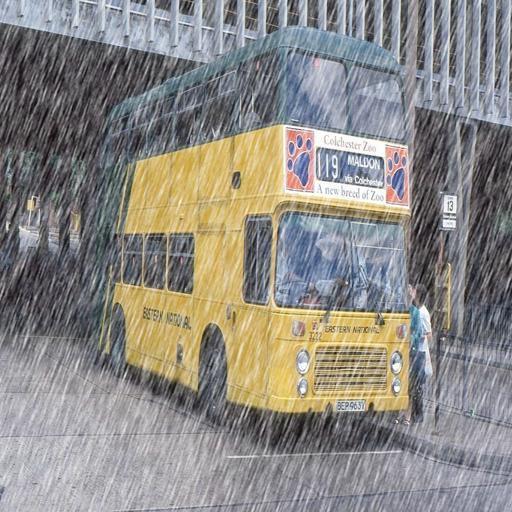}
	\includegraphics[width=2.7cm,height = 1.58cm]{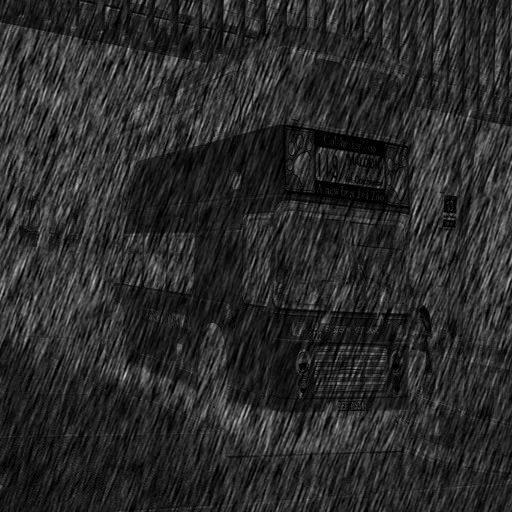}\\
	(a)\hskip50pt(b)\\
	\includegraphics[width=2.7cm,height = 1.58cm]{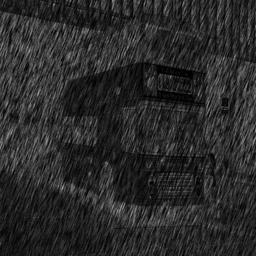}
	\includegraphics[width=2.7cm,height = 1.58cm]{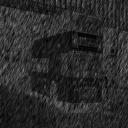}\\
	(c)\hskip50pt(d)\\
	\vskip-10pt
	\caption{(a) Input rainy image, $y$.  (b), (c), and (d) are the residual maps $r_{\x1},\; r_{\x2},\;r_{\x4}$ at scales 1.0, 0.5 and 0.25, respectively.  Note that the residual maps at different scales have the same direction and density.}
	\label{Fig:residual}
	\vskip-15pt
\end{figure}

\begin{figure}[htp!]
	\centering
	\includegraphics[width=2.25cm, height=4.5cm]{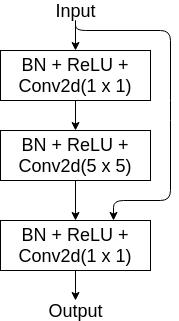}\hskip20pt
	\includegraphics[width=1.7cm, height=4.5cm]{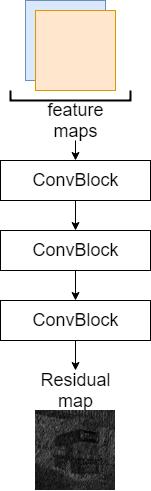}\hskip20pt
	\includegraphics[width=2.25cm, height=4.5cm]{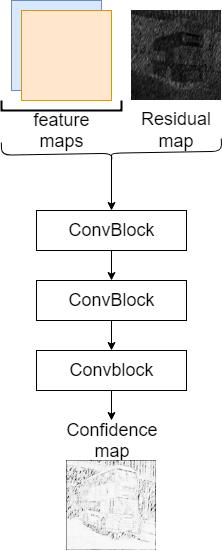}\\
	(a)\hskip70pt(b)\hskip70pt(c)\\
	\vskip-10pt
	\caption{(a) Convolutional block (ConvBlock).  BN - batchnormalization, ReLU - Rectified Linear Units, Conv2d($m\times m$) - convolutional layer with kernel of size $m\times m$. (b) Residual Network (RN). (c) Confidence map Network (CN).}
	\label{Fig:blocks}
	\vskip-10pt
\end{figure}

Let $ r_{\x2} $ (0.5 scale size of $r$) and $r_{\x4}$ (0.25 scale size of $r$) be the residual maps at different scales.   As can be seen from Figure \ref{Fig:residual}, the residual maps $r_{\x1},\; r_{\x2},$ and $r_{\x4}$ have the same direction and density at each location of the image. To estimate these residual maps, we start with the Unet architecture \cite{Authors15b} as the base network. We use the convolutional block (ConvBlock as shown in Figure \ref{Fig:blocks}(a)) as the building block of our base network.  The base network can be described as follows:\\
\vskip-10pt
\noindent ConvBlock(3,32)-AvgPool-ConvBlock(32,32)-AvgPool-Convblock(32,32)-AvgPool-ConvBlock(32,32)-AvgPool-ConvBlock(32,32)-UpSample-ConvBlock(64,32)-UpSample-ConvBlock(67,32)-UpSample-ConvBlock(67,16)-ConvBlock(16,16)-Conv2d($3\times 3$),\\
\vskip-10pt
\noindent where AvgPool is the average pooling layer, UpSample is the upsampling convolution layer, and ConvBlock$(i,j)$ indicates ConvBlock with $i$ input channels and $j$ output channels.  A Refinement Network(RFN) is used at the end of Unet to produce de-rained image.  The Refinement Network(RFN) consists of the following blocks\\ 
\vskip-10pt
\noindent Conv2d($7\times 7$)-Conv2d($3\times 3$)-tanh(),\\
\vskip-10pt
\noindent which takes $y-\hat{r}_{i}$ as the input and generates $\hat{x}_{i}$ (i.e. de-rained image) as the output.  Here,  Conv2d($m\times m$) represents 2D convolution using the kernel of size $m\times m$.

\begin{figure}[htp!]
	\centering
	\includegraphics[width=2.7cm,height = 1.58cm]{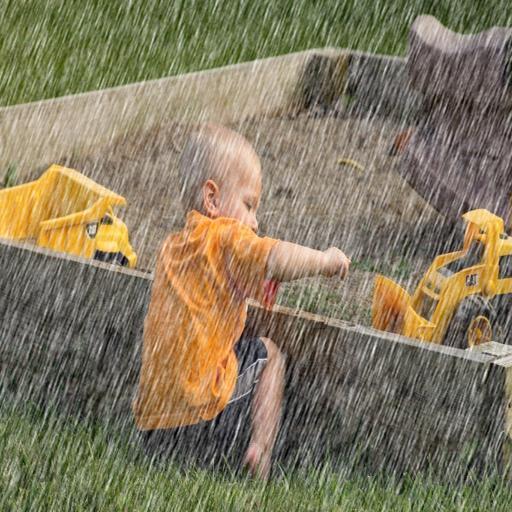}
	\includegraphics[width=2.7cm,height = 1.58cm]{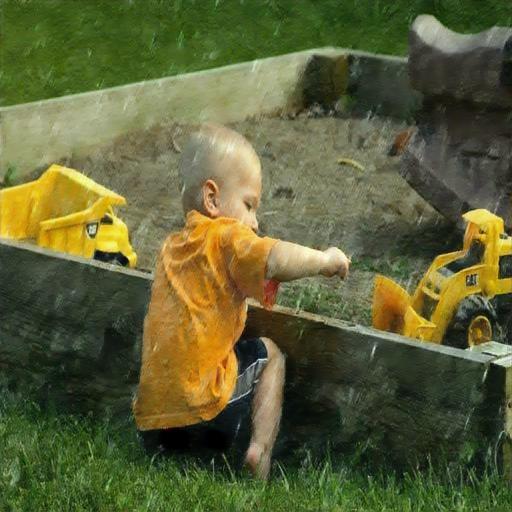}
	\includegraphics[width=2.7cm,height = 1.58cm]{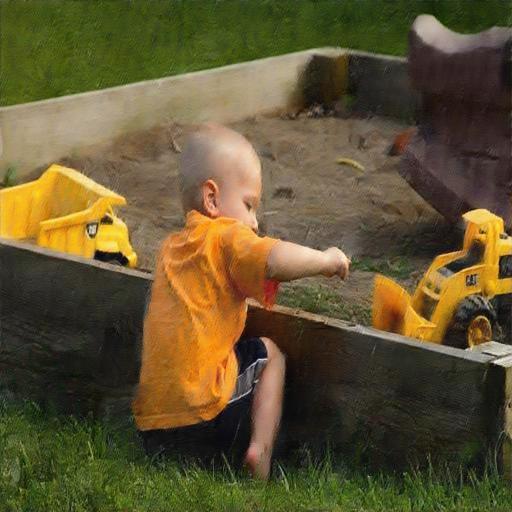}\\
	(a)\hskip60pt(b)\hskip60pt(c)\\
	\includegraphics[width=2.7cm,height = 1.58cm]{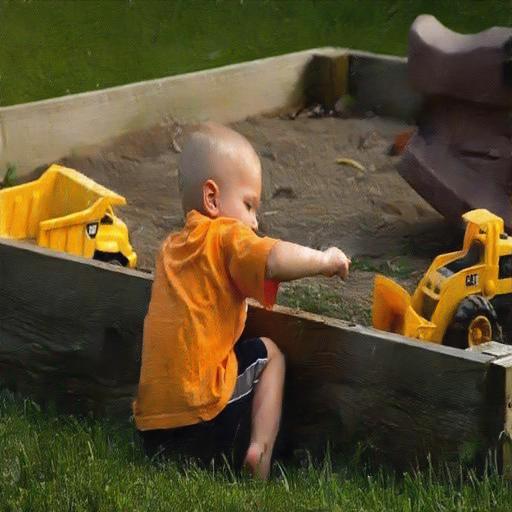}
	\includegraphics[width=2.7cm,height = 1.58cm]{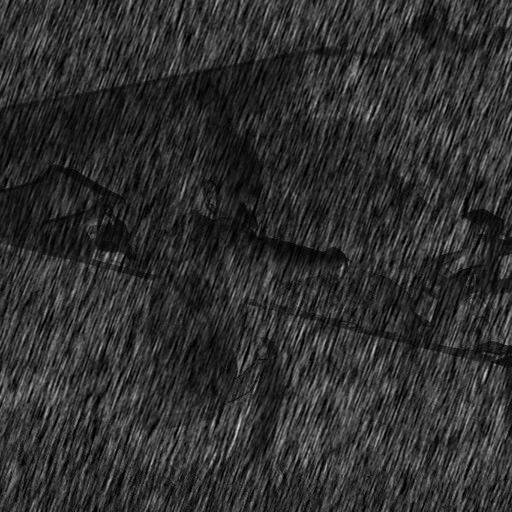}
	\includegraphics[width=2.7cm,height = 1.58cm]{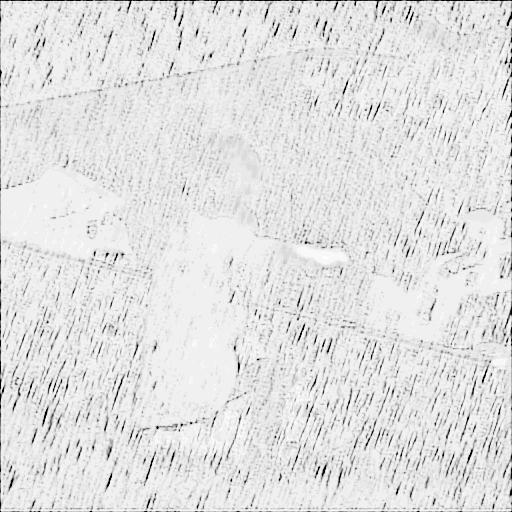}\\
	(d)\hskip60pt(e)\hskip60pt(f)\\
	\vskip-10pt
	\caption{(a) Input rainy image, $y$.  (b) De-rained image using the base network. (c) De-rained using \cite{Authors18}. (d) De-rained using the proposed UMRL  method.  (e) The residual map.  (f) The confidence map at scale 1.0($\x1$).}
	\label{Fig:exp1}
	\vskip-15pt
\end{figure}

\subsection{UMRL Network}
Rainy streaks are high frequency components and existing de-raining methods either tend to remove high frequencies that are not rain streaks or do not remove the rain near high frequency components of the clean image like edges as shown in the Figure \ref{Fig:exp1}. To address this issue, one can use the information about the location in image where network might go wrong in estimating the residual value.  This can be done by estimating a confidence value corresponding to the estimated residual value and guide the network to remove  the artifacts, especially near the edges. For example, we can observe clearly from Figure \ref{Fig:exp1} that the residual map and its corresponding confidence map were able to capture the regions where there is high probability of incorrect estimates. We estimate the residual value and its corresponding confidence map at different scales (1.0($\x1$), 0.5($\x2$) and 0.25($\x4$)) of the input size.  This information is then fed back to the subsequent layers so that the network can learn the residual value at each location, given the computed residual value and confidence value at a lower scale.

\subsubsection{Residual and Confidence Map Networks}
Feature maps at different scales such as  $\x2$ and $\x4$ are given as input to the Residual Network (RN) to estimate the residual map at the corresponding scale as shown in the Figure \ref{Fig:UMRL}. RN consists of the  following sequence of convolutional layers,\\
\vskip-10pt
\noindent  Convblock(64,32)-Convblock(32,32)-Convblock(32,3)\\
\vskip-10pt
\noindent   as shown in Figure \ref{Fig:blocks}(b). We use the estimated residual map and the feature maps as input to the Confidence map Network (CN) to compute the confidence measure at every pixel, which indicates how sure the network is about the residual value at each pixel. CN consists of the following sequence of convolutional layers,\\
\vskip-10pt
\noindent Convblock(67,16)-Convblock(16,16)-Convblock(16,3)\\ 
\vskip-10pt
\noindent as shown in the Figure \ref{Fig:blocks}(c). Given the estimated residual map and the corresponding feature maps as input to the confidence map network, it estimates $c_{\x4}$ and $c_{\x2}$. The element wise product of $\hat{r_i}$ and $c_i$ is computed, and up-sampled to pass it as an input to the subsequent layer of the UMRL network as shown in Figure \ref{Fig:UMRL} for $i\in \{\x2,\x4\}$. Given the output residual map $r_{\x1}$ and the feature maps of the final layer of UMRL as input to CN, we get $c_{\x1}$. We compute the de-rained image at different scale as 
\begin{equation}
\hat{x_i} = \text{RFN} (y_i - \hat{r_i} ),
\end{equation}
where RFN is the Refinement Network, $y_i$ and $\hat{x_i}$ are the input rainy image and the output de-rained image at scales, $i \in \{\x1 (1.0),\x2 (0.5),\x4 (0.25)\}$.  We use the confidence guided loss and the preceptual loss to train our network.

\subsubsection{Loss for UMRL}
We use the confidence to guide the residual learning in the training stage of UMRL network. We define the confidence guided loss as,
\begin{equation}
\begin{aligned}
\mathcal{L}_l &= \sum_{i \in \{\x1,\x2,\x4\}}\lVert(c_i\odot\hat{x}_i)-(c_i\odot x_i)\rVert_1, \\
\mathcal{L}_c &= \sum_{i \in \{\x1,\x2,\x4\}} \big(\sum_j\sum_k \log(c_{i_{jk}})\big),\\
\mathcal{L}_u &= \mathcal{L}_l - \lambda_1 \mathcal{L}_c,
\end{aligned}
\end{equation}
where $\odot$ is the element wise product. Here, $\mathcal{L}_l$ tries to minimize the L1-norm between $\hat{x}_i$ and $x_i$ and also the value of $c_{i_{jk}}$.  On the other hand, $\mathcal{L}_c$ tries to increase $c_{i_{jk}}$ by making it close to 1. A trivial solution for $\mathcal{L}_l$ can be seen as $c_{i_{jk}}=0\; \forall\:i,\:j,\:k$. To avoid this, we construct $\mathcal{L}_u$ as a linear combination of $\mathcal{L}_l$ and $\mathcal{L}_c$, where $\mathcal{L}_c$ acts as a regularizer to avoid the trivial solution. Similar loss has been used for classification and regression tasks in methods \cite{KendallGal2017UncertaintiesB,kendall2017multi}. However, to the best of our knowledge ours is the first attempt to use this kind of loss in image restoration tasks. Inspired by the importance of the perceptual loss in many image restoration tasks \cite{Authors16b,Authors17g}, we use it to further improve the visual quality of the de-rained images. The perceptual loss is feature based loss, and in our case, extracted features  from layer $relu1\_2$ of pretrained network VGG-16\cite{Authors14c}, and computed perceptual loss similar to method proposed in \cite{Johnson2016Perceptual,zhang2017multistyle}. Let $F(.)$ denote the features obtained using the VGG16 model \cite{Authors14c}, then the perceptual loss is defined as follows
\begin{equation}
\mathcal{L}_p = \frac{1}{NHW}\sum_i \sum_j \sum_k \lVert F(\hat{x}_1)^{i,j,k}-F(x_1)^{i,j,k}\rVert_2^{2},
\end{equation}
where $N$ is the number of channels of $F(.)$, $H$ is the height and $W$ is the width of feature maps. The overall loss used to train the UMRL network is,
\begin{equation}
\mathcal{L} = \mathcal{L}_l - \lambda_1 \mathcal{L}_c + \lambda_2 \mathcal{L}_p,
\end{equation}
where $\lambda_1$ and $\lambda_2$ are two parameters.

\begin{figure}[htp!]
	\begin{center}
		\centering
		\includegraphics[width=8cm]{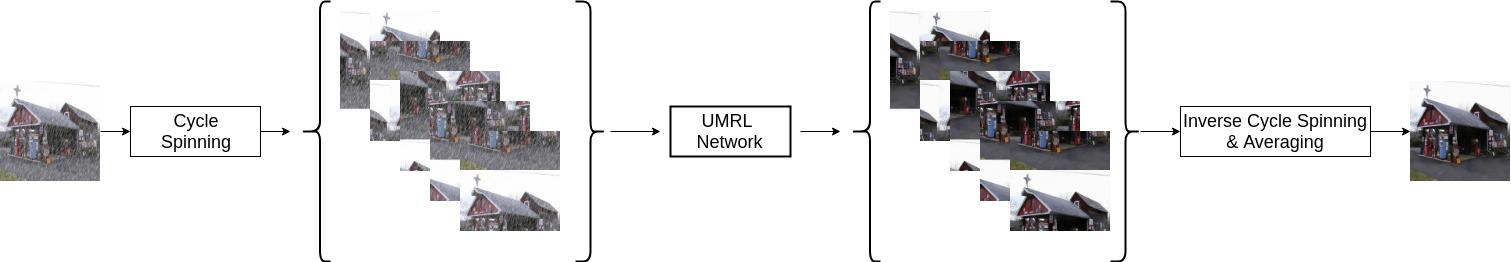}\\
		\vskip -10pt	\caption{The idea behind cycle spinning using the UMRL network.}
		\label{Fig:Cyc}
	\end{center}
	\vskip-20pt
\end{figure}

\subsection{Cycle Spinning}
\begin{figure}[htp!]
	\centering
	\includegraphics[width=2.7cm,height = 1.58cm]{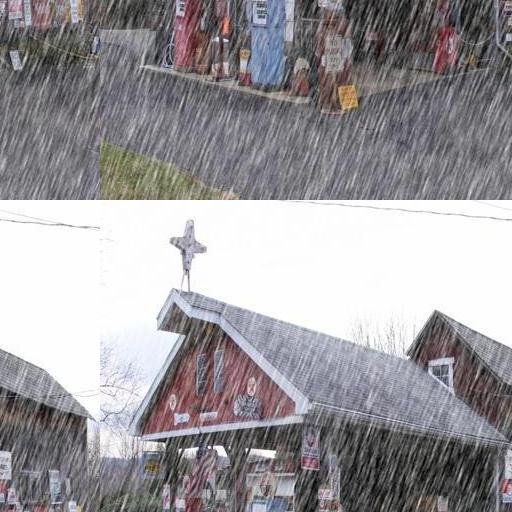}
	\includegraphics[width=2.7cm,height = 1.58cm]{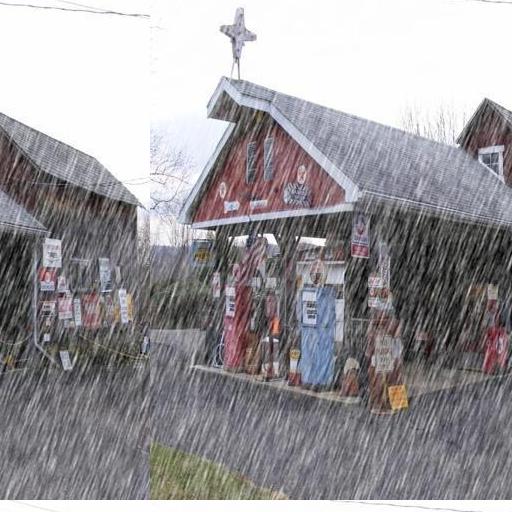}
	\includegraphics[width=2.7cm,height = 1.58cm]{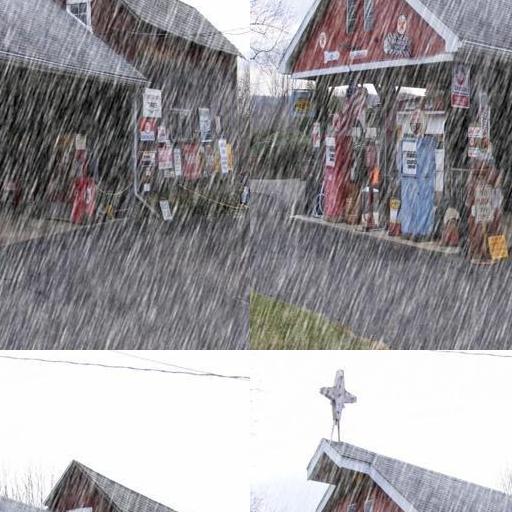}\\
	(a)\hskip60pt(b)\hskip60pt(c)\\
	\includegraphics[width=2.7cm,height = 1.58cm]{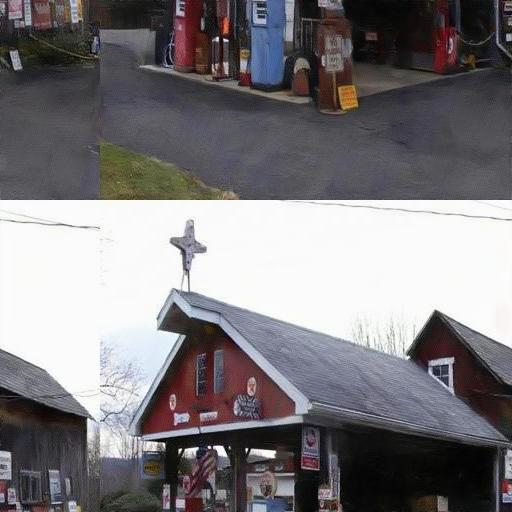}
	\includegraphics[width=2.7cm,height = 1.58cm]{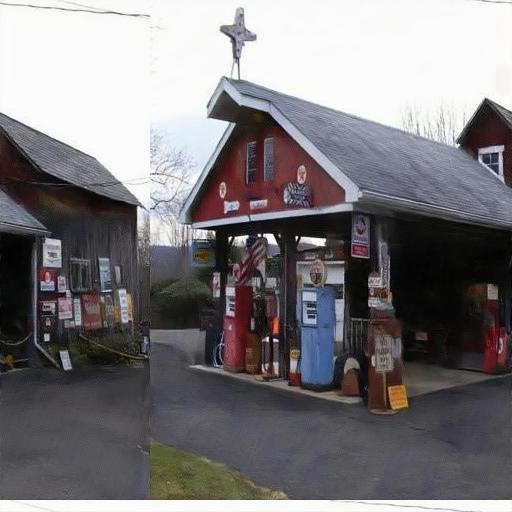}
	\includegraphics[width=2.7cm,height = 1.58cm]{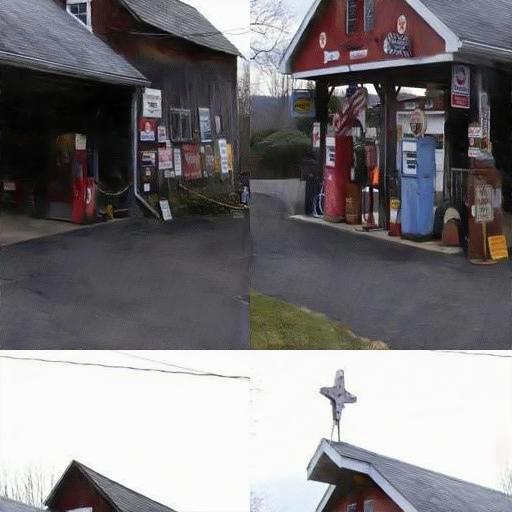}\\
	(d)\hskip60pt(e)\hskip60pt(f)\\
	\vskip-10pt
	\caption{Cyclically spinned images with (a) $p=100$, $q=200$, (b) $p=0$, $q=200$, and (c) $p=300$, $q=400$.  (d),(e),(f) are the  corresponding de-rained images using UMRL.}
	\label{Fig:exp2}
	\vskip-10pt
\end{figure}

As discussed earlier, cycle spinning was originally proposed to minimize the artifacts near the edges introduced by the orthogonal wavelets when de-noising images \cite{Authors95}.  In this work, we adapt this idea to further improve the de-raining performance of UMRL.  Figure~\ref{Fig:Cyc} gives an overview of cycle spinning using UMRL.  Let $T_{cs}(.,p,q)$ be the function to shift an image cyclically by $p$ rows and $q$ columns. Given an image of size $m \times n$, we shift the image cyclically in steps of $p$ rows and $q$ columns to get the shifted images as shown in the Figure \ref{Fig:Cyc}. We then de-rain the shifted images using the UMRL network, inverse shift and average them to get the final de-rained image during testing. Figure~\ref{Fig:exp2} shows an example of cyclically spun input images and the corresponding de-rained images.  By applying cycle spinning to our method, we are able to remove some artifacts introduced by the original UMRL network.  In particular, as will be shown later, cycle spinning can be applied to any CNN-based de-raining method to further improve its performance.

\section{Experimental Results}
In this section, we evaluate the performance of our method on both synthetic and real images.  Peak-Signal-to-Noise Ratio (PSNR) and Structural Similarity index (SSIM) \cite{SSIM} measures are used to compare the performance of different methods on synthetic images.   We visually inspect the performance of different methods on real images, as we don't have the ground truth clean images. The performance of the proposed UMRL method is compared against several recent state-of-the-art algorithms such as (a) Gaussian mixture model (GMM) based \cite{Authors16} (CVPR’16) (b) Fu et al.\cite{Authors17d} CNN method (TIP'17), (c) Joint Rain Detection and Removal (JORDER) \cite{Authors17h}(CVPR’17), (d) Deep detailed Network (DDN)\cite{Authors17f} (CVPR'17) (e) Zhu et al. \cite{Authors17c} (JBO) (ICCV’17) (f) Density-aware Image De-raining method using a Multistream
Dense Network (DID-MDN) \cite{Authors18} (CVPR'18).

\begin{figure}[htp!]
	\centering
	\begin{minipage}{.155\textwidth}
		\centering
		\caption*{\emph{\footnotesize{PSNR:15.3 SSIM: 0.71}}}
		\vskip-10pt
		\includegraphics[width=2.7cm,height = 1.8cm]{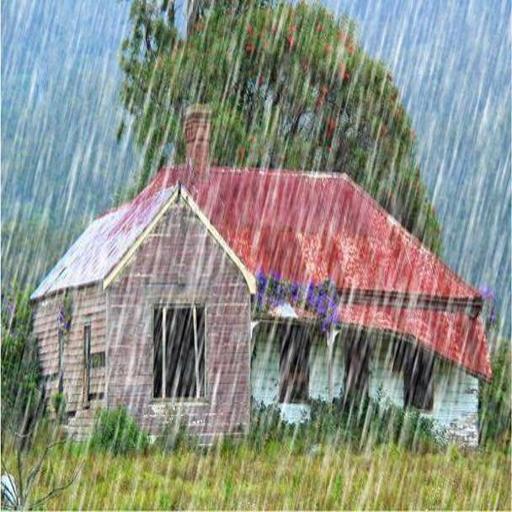}
		\captionsetup{labelformat=empty}
		\captionsetup{justification=centering}
	\end{minipage}
	\begin{minipage}{.155\textwidth}
		\centering
		\caption*{\emph{\footnotesize{PSNR:24.5 SSIM: 0.87}}}
		\vskip-10pt
		\includegraphics[width=2.7cm,height = 1.8cm]{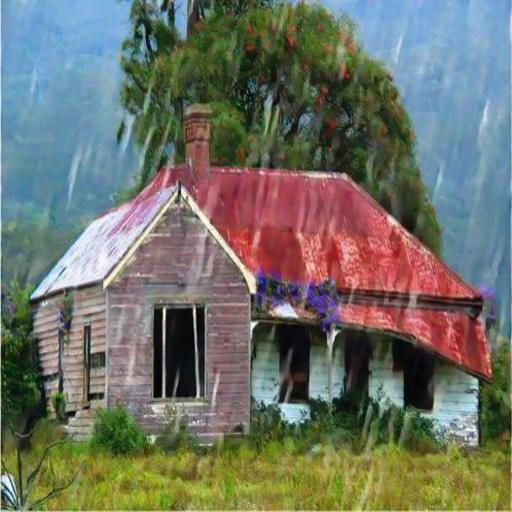}
		\captionsetup{labelformat=empty}
		\captionsetup{justification=centering}
	\end{minipage}
	\begin{minipage}{.155\textwidth}
		\centering
		\caption*{\emph{\footnotesize{PSNR:26.9 SSIM: 0.92}}}
		\vskip-10pt
		\includegraphics[width=2.7cm,height = 1.8cm]{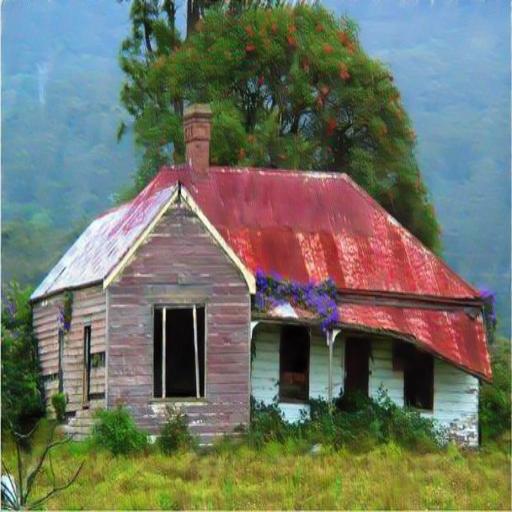}
		\captionsetup{labelformat=empty}
		\captionsetup{justification=centering}
	\end{minipage}\\
	(a)\hskip70pt(b)\hskip70pt(c)\\
	\includegraphics[width=2.7cm,height = 1.8cm]{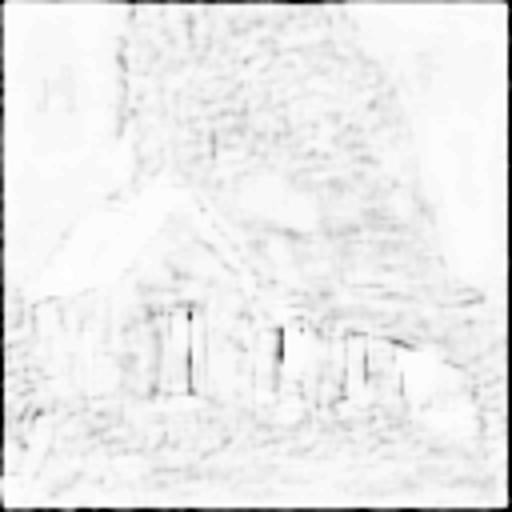}
	\includegraphics[width=2.7cm,height = 1.8cm]{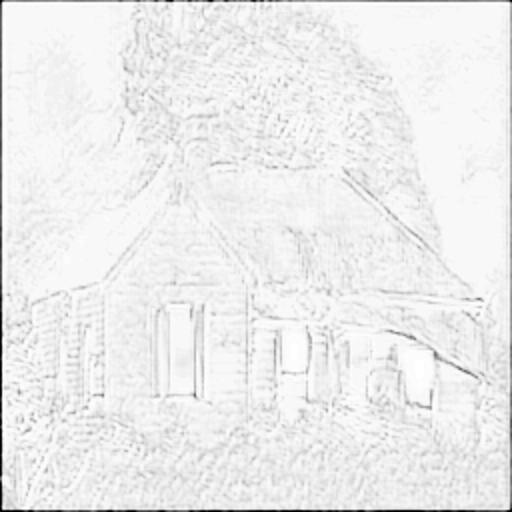}
	\includegraphics[width=2.7cm,height = 1.8cm]{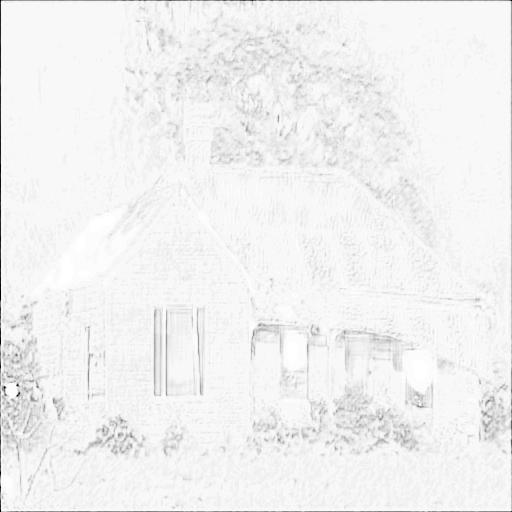}\\
	(d)\hskip70pt(e)\hskip70pt(f)\\
	\includegraphics[width=2.7cm,height = 1.8cm]{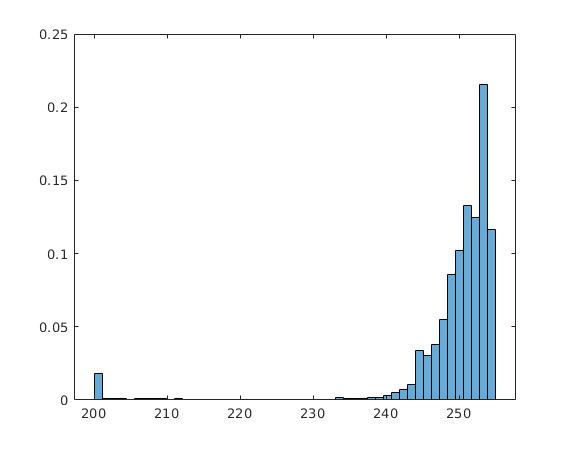}
	\includegraphics[width=2.7cm,height = 1.8cm]{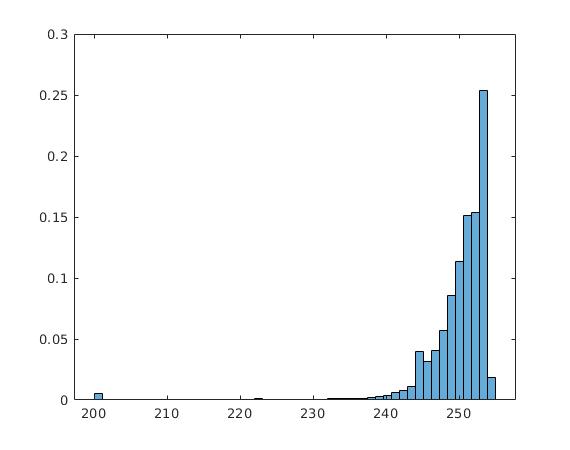}
	\includegraphics[width=2.7cm,height = 1.8cm]{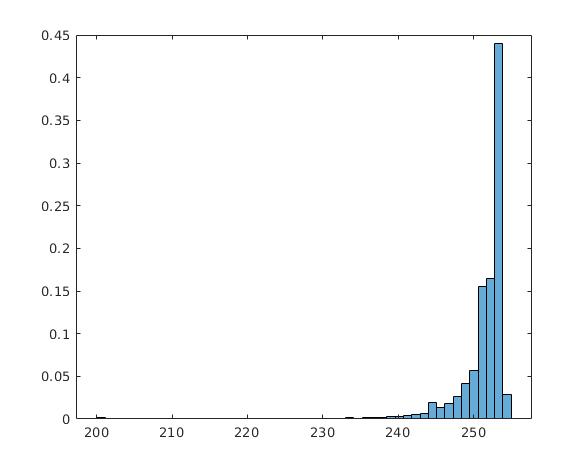}\\
	(g)\hskip70pt(h)\hskip70pt(i)\\
	\vskip-10pt
	\caption{(a) Input rainy image. (b) De-rained image using BN + RN. (c) De-rained using BN + RN + CN (UMRL). (d),(e), and (f) are the corresponding confidence maps at scales {$\x4,\x2,\x1$}. (g),(h), and (i) are the corresponding normalized histograms, that is sum of all bin values is equal to 1.}
	\label{Fig:exp3}
	\vskip-20pt
\end{figure}

\subsection{Training and Testing Details}
The UMRL network is trained using the synthetic image datasets created by the authors of \cite{Authors18,Authors17e}.   The dataset in  \cite{Authors18} consists of 12000 images with different rain levels like low, medium and high.  The dataset in \cite{Authors17e} contains 700 training images. The $(y, x)$ rainy-clean image pairs are shifted randomly $p$ rows and $q$ columns using $T_{cs}(.,p,q)$ to obtain $y_s$, $x_s$, respectively. The shifted pairs ($y_s, x_s$) are used to train UMRL using the loss $\mathcal{L}$. The Adam optimizer with the batch size of 1 is used to train the network. Learning rate is set to 0.001 for first 10 epochs and 0.0001 for the remaining epochs. During training initially $\lambda_1$ and $\lambda_2$ are set equal to 0.1 and 1.0, respectively, but when the mean of all values in the confidences maps $c_{\x1},\;c_{\x2}$ and $c_{\x4}$ is greater than 0.8 then $\lambda_1$ is set equal to 0.03. UMRL is trained for 30 epochs that is a total of $30 \x 12700$ iterations.

Similar to the previous approaches \cite{Authors18}, we evaluate the performance of UMRL using the datasets \textit{Test-1} containing 1200 images from \cite{Authors18}, and  \textit{Test-2} containing 1000 images from \cite{Authors17d}. We use the real-world rainy images provided by Zhang et al.\cite{Authors17e} and Yang et al. \cite{Authors17h} for testing UMRL based cycle spinning method. The testing images are cyclically shifted in steps of 50 rows and 50 columns using $T_{cs}(.,.,.)$ and fed as input to UMRL for de-raining, further these de-rained images are inverse shifted and averaged to get the final output.

\begin{figure}[h!]
	\centering
	\begin{minipage}{.23\textwidth}
		\centering
		\caption*{\emph{\footnotesize{PSNR:23.01 SSIM: 0.81}}}
		\vskip-10pt
		\includegraphics[width=3cm,height=1.75cm]{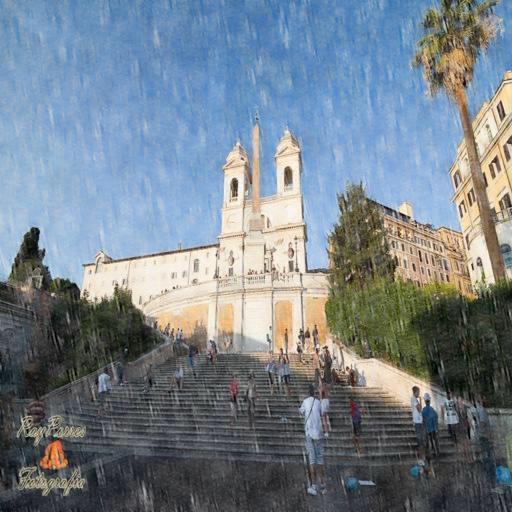}
		\captionsetup{labelformat=empty}
		\captionsetup{justification=centering}
	\end{minipage}
	\begin{minipage}{.23\textwidth}
		\centering
		\caption*{\emph{\footnotesize{PSNR:25.69 SSIM: 0.88}}}
		\vskip-10pt
		\includegraphics[width=3cm,height=1.75cm]{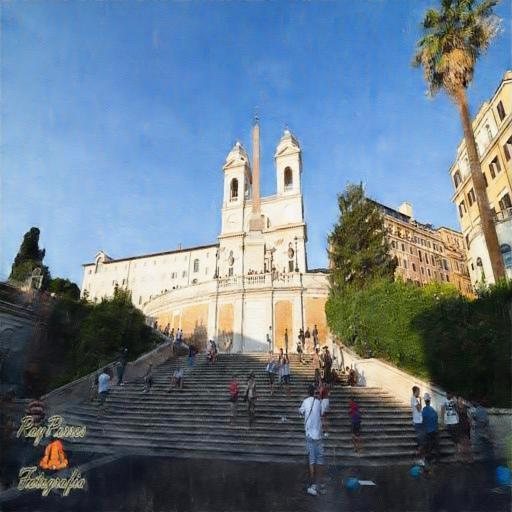}
		\captionsetup{labelformat=empty}
		\captionsetup{justification=centering}
	\end{minipage}\\
	(a)\hskip100pt(b)\\
	\begin{minipage}{.23\textwidth}
		\centering
		\caption*{\emph{\footnotesize{PSNR:26.31 SSIM: 0.90}}}
		\vskip-10pt
		\includegraphics[width=3cm,height=1.75cm]{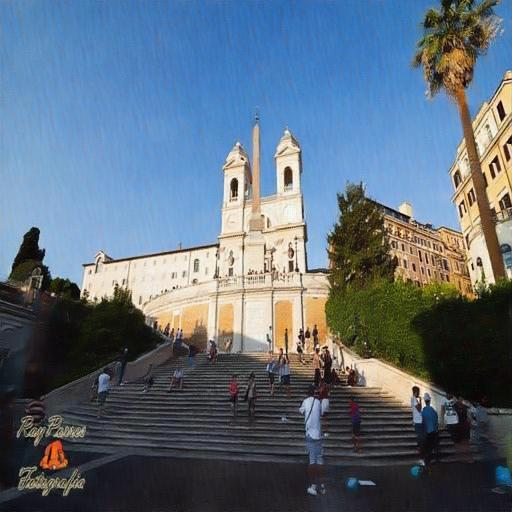}
		\captionsetup{labelformat=empty}
		\captionsetup{justification=centering}
	\end{minipage}
	\begin{minipage}{.23\textwidth}
		\centering
		\caption*{\emph{\footnotesize{PSNR:27.10 SSIM: 0.92}}}
		\vskip-10pt
		\includegraphics[width=3cm,height=1.75cm]{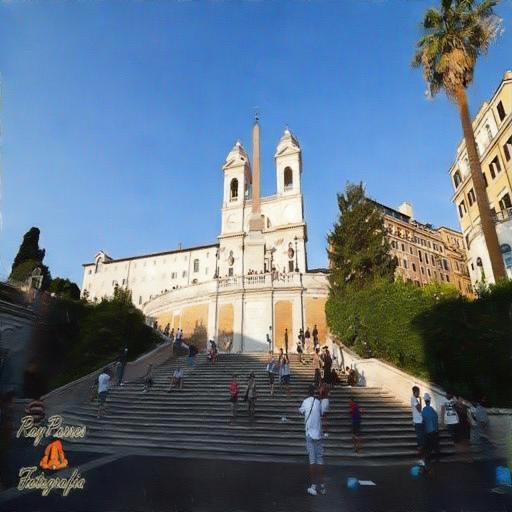}
		\captionsetup{labelformat=empty}
		\captionsetup{justification=centering}
	\end{minipage}\\
	(c)\hskip100pt(d)\\
	\vskip-10pt
	\caption{De-rained images using (a) DDN \cite{Authors17f}, (b) DID-MDN \cite{Authors18}, (c) UMRL, and (d) UMRL + cycle spinning.}
	\label{Fig:exp5}
	\vskip-20pt
\end{figure}

\subsection{Ablation Study}
We study the performance of each block's contribution to the UMRL network by conducting extensive experiments on the test datasets. We start with the Unet-based base network (BN) and then add one component at a time to see the significance each component brings to the network in estimating the final de-rained image. Table \ref{ablation}, shows the contribution of each block on the UMRL network.  Note that BN and BN+RN are trained using a linear combination of L1-norm and $\mathcal{L}_p$ as loss (L1$+\mathcal{L}_p$).   The UMRL is trained using the overall loss, $\mathcal{L}$. It can be seen from Table \ref{ablation} that as more components (i.e RN and CN) are being added to the base network, the performance improves significantly.  The base network, BN itself produces poor results.  However, when RN is added to BN, the performance improves significantly.  In particular, BN+RN is already able to produce results that are comparable to DID-MDN \cite{Authors18}.  The combination of BN, RN and CN (i.e UMRL) produces the best results.  Furthermore, by comparing the last two columns of Table \ref{ablation} we see that cycle spinning further improves the performance of UMRL.   Using cycle spinning, we are able to gain the performance improvement of approximately $0.3$ dB on both datasets as it was able to remove the artifacts near edges. From the Figure \ref{Fig:exp5} by zooming-in, we can clearly observe the cycle spinning is helping the method to remove small rain streaks in the sky and on the edges of building.

\begin{table}[htp!]
	\vskip-10pt
	\caption{PSNR and SSIM (PSNR$|$SSIM) results corresponding to the ablation study.}
	\vskip-10pt
	\resizebox{8.25cm}{!}{
		\label{ablation}
		\begin{tabular}{|c|c|c|c|c|c|c|}
			\hline
			Dataset & \begin{tabular}[c]{@{}c@{}}Rainy\\ Image\end{tabular} & DID-MDN \cite{Authors18}& BN & BN+RN & BN+RN+CN (UMRL) & \begin{tabular}[c]{@{}c@{}}UMRL+\\ cycle spinning\end{tabular} \\ \hline
			\textit{Test-1} & 21.15$|$0.77 & 27.95$|$0.91 & 24.25$|$0.83   & 27.65$|$0.87 & 29.42$|$0.91 & \textbf{29.77$|$0.92} \\ \hline
			\textit{Test-2} & 19.31$|$0.77 & 26.08$|$0.90 &  23.32$|$0.83  &  25.88$|$0.87 & 26.47$|$0.91 & \textbf{26.67$|$0.92} \\ \hline
		\end{tabular}
	}
	\vskip-10pt
\end{table}

\begin{figure*}[htp!]
	\centering
	\begin{minipage}{.16\textwidth}
		\centering
		\caption*{\emph{PSNR: 18.75\\SSIM: 0.67}}
		\vskip-10pt
		\includegraphics[width=2.7cm,height=1.58cm]{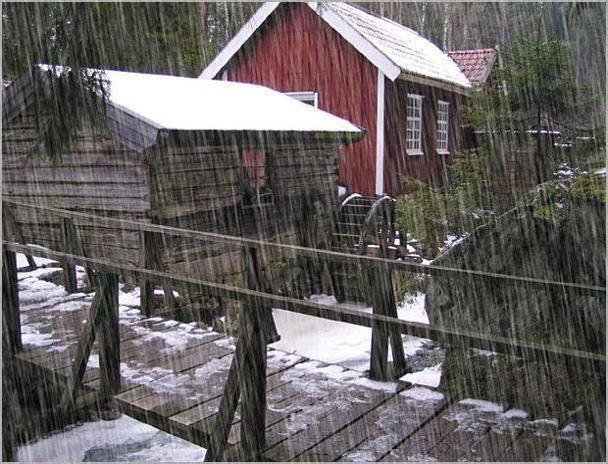}
		\captionsetup{labelformat=empty}
		\captionsetup{justification=centering}
	\end{minipage}  
	\begin{minipage}{.16\textwidth}
		\centering
		\caption*{\emph{PSNR:20.02 \\SSIM: 0.74}}
		\vskip-10pt
		\includegraphics[width=2.7cm,height=1.58cm]{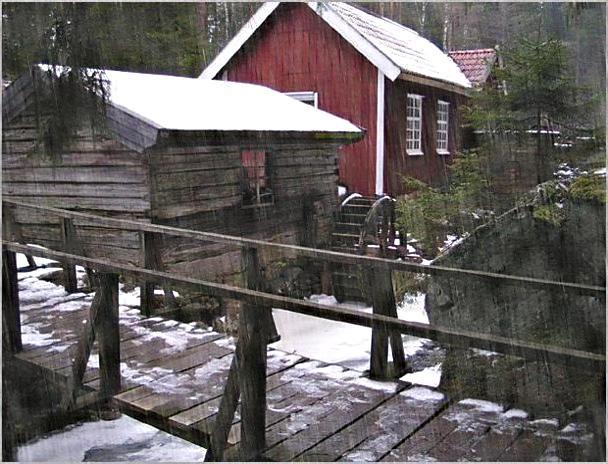}
		\captionsetup{labelformat=empty}
		\captionsetup{justification=centering}
	\end{minipage}
	\begin{minipage}{.16\textwidth}
		\centering
		\caption*{\emph{PSNR: 26.12 \\SSIM:0.82}}
		\vskip-10pt
		\includegraphics[width=2.7cm,height=1.58cm]{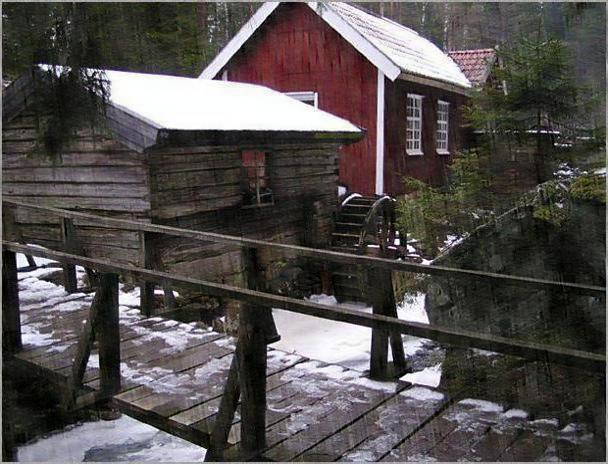}
		\captionsetup{labelformat=empty}
		\captionsetup{justification=centering}
	\end{minipage}
	\begin{minipage}{.16\textwidth}
		\centering
		\caption*{\emph{PSNR: 25.27 \\SSIM: 0.82}}
		\vskip-10pt
		\includegraphics[width=2.7cm,height=1.58cm]{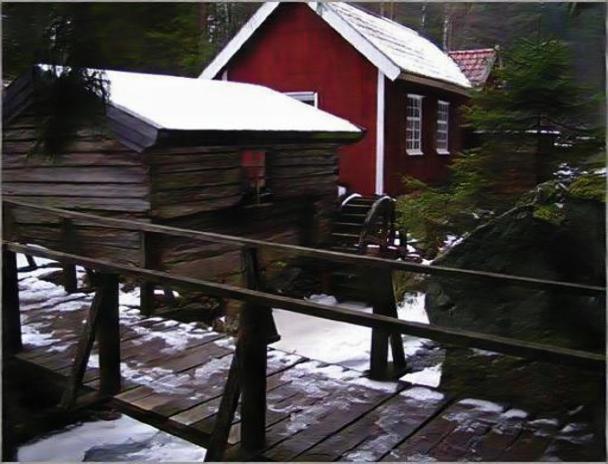}
		\captionsetup{labelformat=empty}
		\captionsetup{justification=centering}
	\end{minipage}
	\begin{minipage}{.16\textwidth}
		\centering
		\caption*{\emph{PSNR: \textbf{27.95} \\SSIM: \textbf{0.87}}}
		\vskip-10pt
		\includegraphics[width=2.7cm,height=1.58cm]{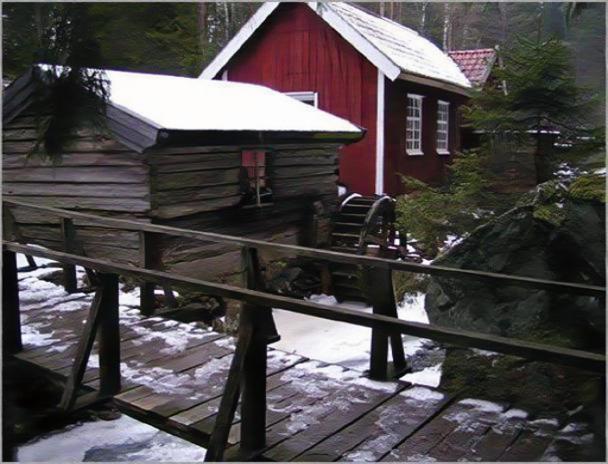}
		\captionsetup{labelformat=empty}
		\captionsetup{justification=centering}
	\end{minipage} 
	\begin{minipage}{.16\textwidth}
		\centering
		\caption*{\emph{PSNR: Inf\\SSIM: 1}}
		\vskip-10pt
		\includegraphics[width=2.7cm,height=1.58cm]{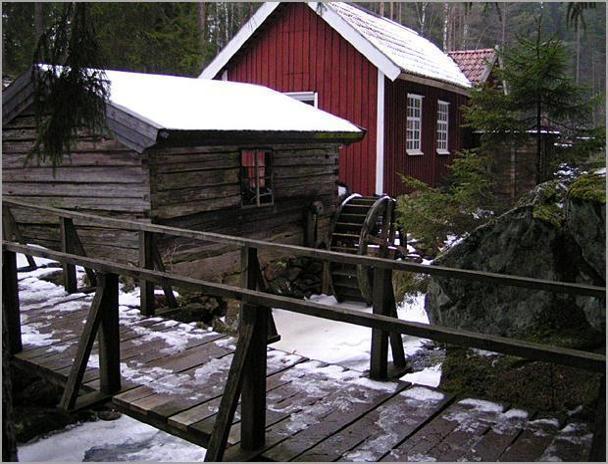}
		\captionsetup{labelformat=empty}
		\captionsetup{justification=centering}
	\end{minipage}\\        \vskip+6pt        
	\begin{minipage}{.16\textwidth}
		\centering
		\caption*{\emph{PSNR: 14.25\\SSIM: 0.59}}
		\vskip-10pt
		\includegraphics[width=2.7cm,height=1.58cm]{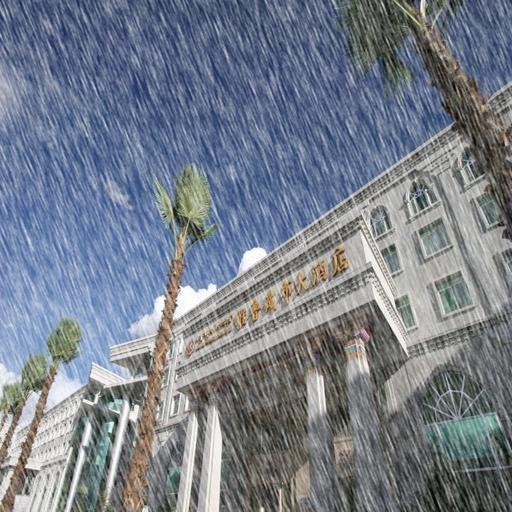}
		\captionsetup{labelformat=empty}
		\captionsetup{justification=centering}
	\end{minipage}  
	\begin{minipage}{.16\textwidth}
		\centering
		\caption*{\emph{PSNR:16.97 \\SSIM: 0.70}}
		\vskip-10pt
		\includegraphics[width=2.7cm,height=1.58cm]{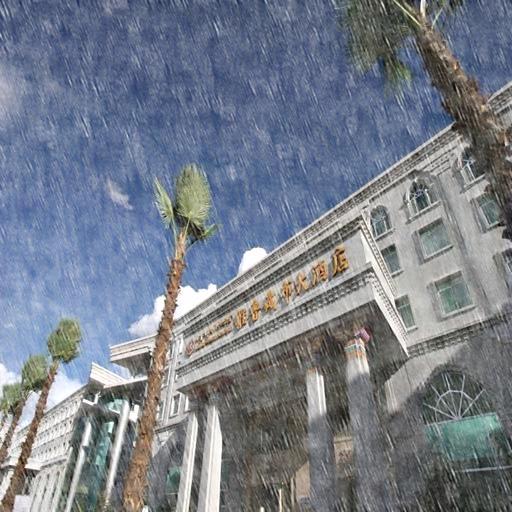}
		\captionsetup{labelformat=empty}
		\captionsetup{justification=centering}
	\end{minipage}
	\begin{minipage}{.16\textwidth}
		\centering
		\caption*{\emph{PSNR: 21.27 \\SSIM:0.78}}
		\vskip-10pt
		\includegraphics[width=2.7cm,height=1.58cm]{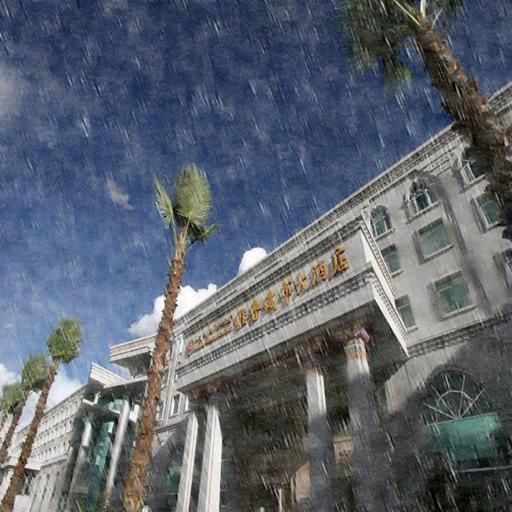}
		\captionsetup{labelformat=empty}
		\captionsetup{justification=centering}
	\end{minipage}
	\begin{minipage}{.16\textwidth}
		\centering
		\caption*{\emph{PSNR: 25.39 \\SSIM: 0.88}}
		\vskip-10pt
		\includegraphics[width=2.7cm,height=1.58cm]{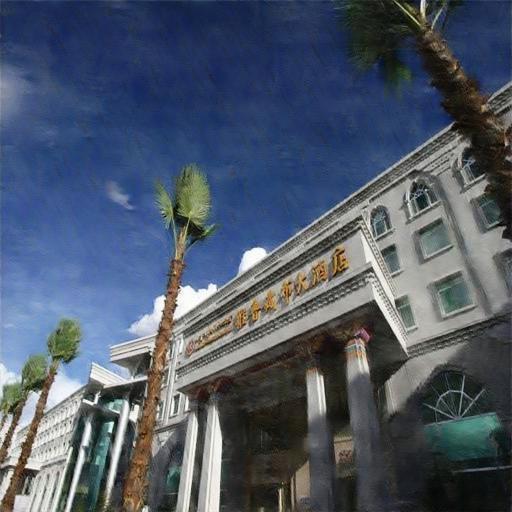}
		\captionsetup{labelformat=empty}
		\captionsetup{justification=centering}
	\end{minipage}
	\begin{minipage}{.16\textwidth}
		\centering
		\caption*{\emph{PSNR: \textbf{26.57} \\SSIM: \textbf{0.9605}}}
		\vskip-10pt
		\includegraphics[width=2.7cm,height=1.58cm]{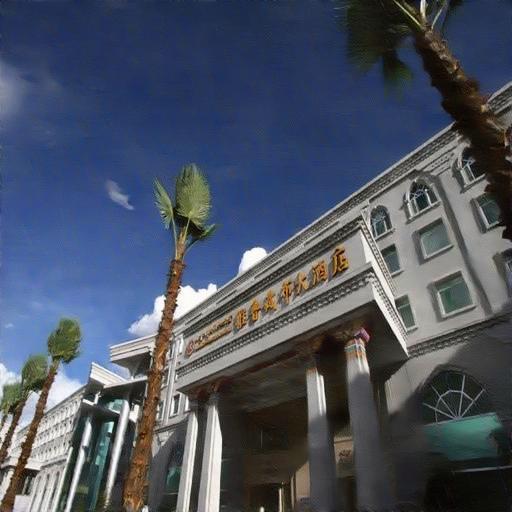}
		\captionsetup{labelformat=empty}
		\captionsetup{justification=centering}
	\end{minipage} 
	\begin{minipage}{.16\textwidth}
		\centering
		\caption*{\emph{PSNR: Inf\\SSIM: 1}}
		\vskip-10pt
		\includegraphics[width=2.7cm,height=1.58cm]{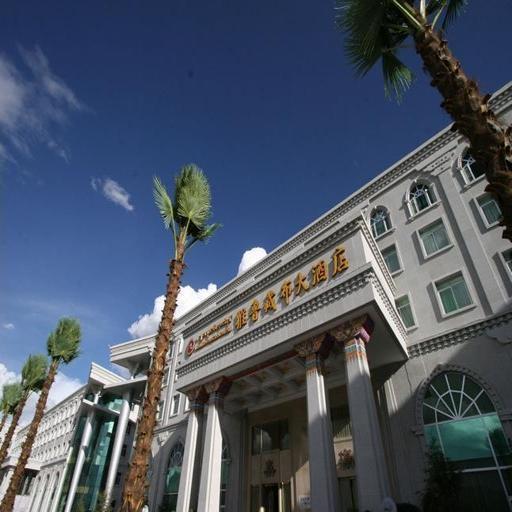}
		\captionsetup{labelformat=empty}
		\captionsetup{justification=centering}
	\end{minipage}\\        \vskip+6pt
	\begin{minipage}{.16\textwidth}
		\centering
		\caption*{\emph{PSNR: 16.26\\SSIM: 0.60}}
		\vskip-10pt
		\includegraphics[width=2.7cm,height=1.58cm]{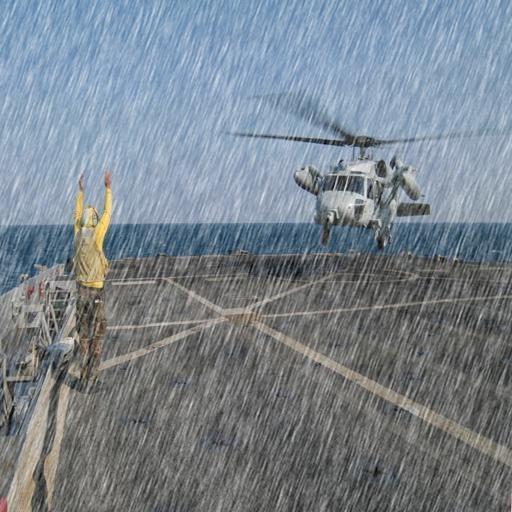}
		\captionsetup{labelformat=empty}
		\captionsetup{justification=centering}
	\end{minipage}  
	\begin{minipage}{.16\textwidth}
		\centering
		\caption*{\emph{PSNR:17.51 \\SSIM: 0.75}}
		\vskip-10pt
		\includegraphics[width=2.7cm,height=1.58cm]{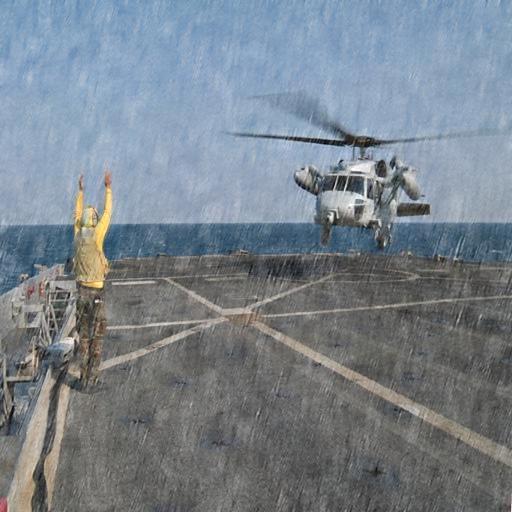}
		\captionsetup{labelformat=empty}
		\captionsetup{justification=centering}
	\end{minipage}
	\begin{minipage}{.16\textwidth}
		\centering
		\caption*{\emph{PSNR: 25.28 \\SSIM:0.83}}
		\vskip-10pt
		\includegraphics[width=2.7cm,height=1.58cm]{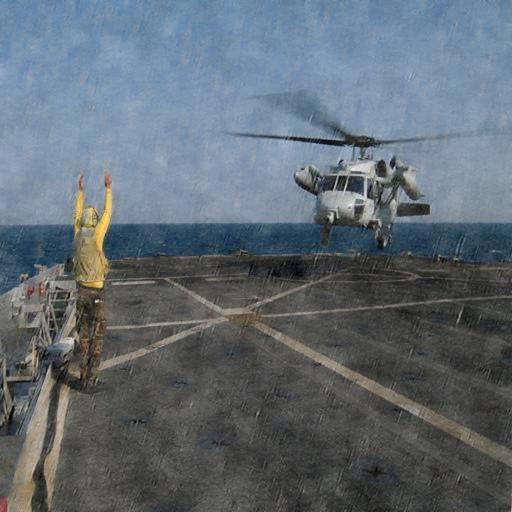}
		\captionsetup{labelformat=empty}
		\captionsetup{justification=centering}
	\end{minipage}
	\begin{minipage}{.16\textwidth}
		\centering
		\caption*{\emph{PSNR: 29.63 \\SSIM: 0.97}}
		\vskip-10pt
		\includegraphics[width=2.7cm,height=1.58cm]{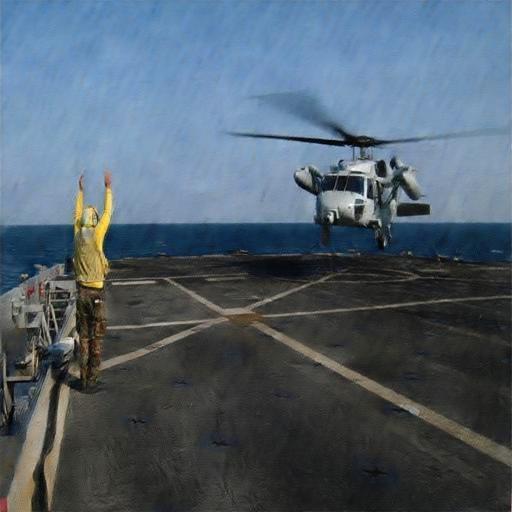}
		\captionsetup{labelformat=empty}
		\captionsetup{justification=centering}
	\end{minipage}
	\begin{minipage}{.16\textwidth}
		\centering
		\caption*{\emph{PSNR: \textbf{30.51} \\SSIM: \textbf{0.98}}}
		\vskip-10pt
		\includegraphics[width=2.7cm,height=1.58cm]{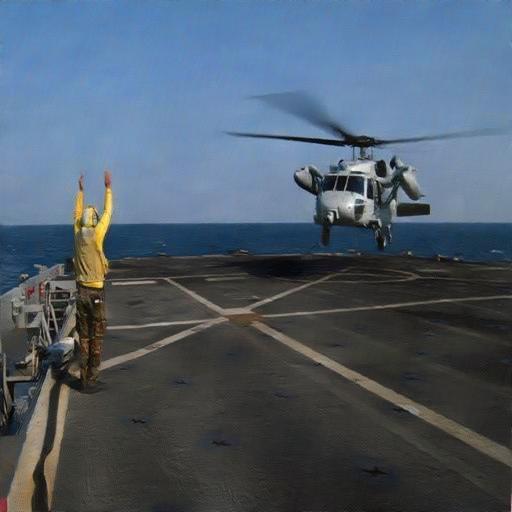}
		\captionsetup{labelformat=empty}
		\captionsetup{justification=centering}
	\end{minipage} 
	\begin{minipage}{.16\textwidth}
		\centering
		\caption*{\emph{PSNR: Inf\\SSIM: 1}}
		\vskip-10pt
		\includegraphics[width=2.7cm,height=1.58cm]{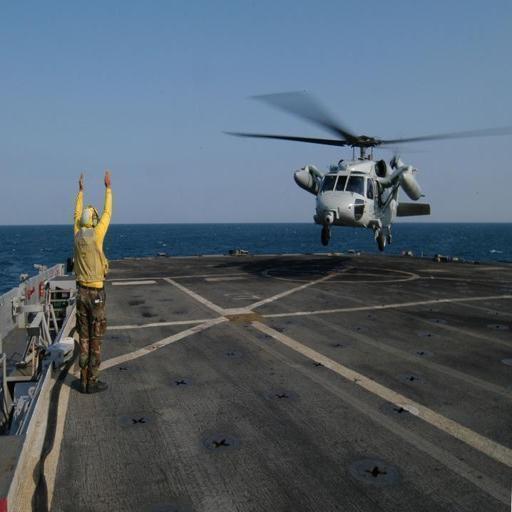}
		\captionsetup{labelformat=empty}
		\captionsetup{justification=centering}
	\end{minipage}\\        \vskip+6pt
	\begin{minipage}{.16\textwidth}
		\centering
		\caption*{\emph{PSNR: 15.58\\SSIM: 0.68}}
		\vskip-10pt
		\includegraphics[width=2.7cm,height=1.58cm]{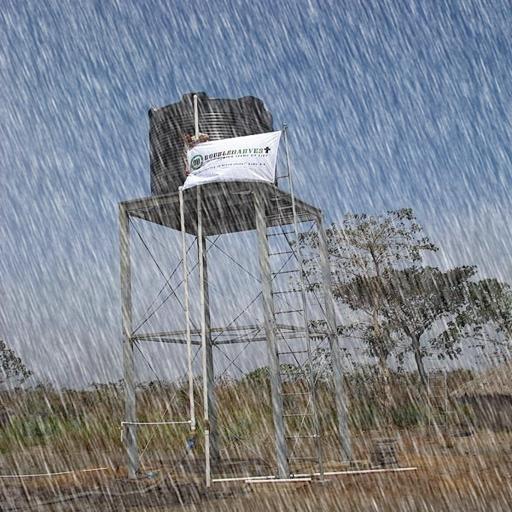}
		\captionsetup{labelformat=empty}
		\captionsetup{justification=centering}
	\end{minipage}  
	\begin{minipage}{.16\textwidth}
		\centering
		\caption*{\emph{PSNR:16.54 \\SSIM: 0.78}}
		\vskip-10pt
		\includegraphics[width=2.7cm,height=1.58cm]{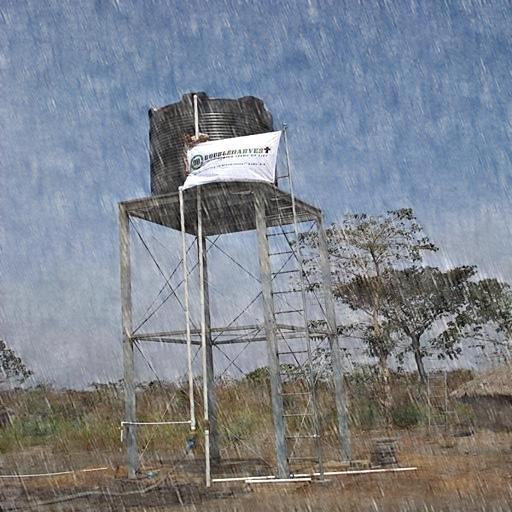}
		\captionsetup{labelformat=empty}
		\captionsetup{justification=centering}
	\end{minipage}
	\begin{minipage}{.16\textwidth}
		\centering
		\caption*{\emph{ PSNR: 23.12 \\SSIM:0.84}}
		\vskip-10pt
		\includegraphics[width=2.7cm,height=1.58cm]{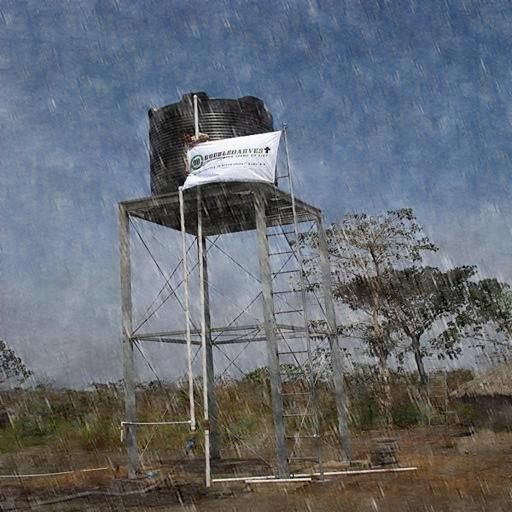}
		\captionsetup{labelformat=empty}
		\captionsetup{justification=centering}
	\end{minipage}
	\begin{minipage}{.16\textwidth}
		\centering
		\caption*{\emph{ PSNR: 25.25 \\SSIM: 0.90}}
		\vskip-10pt
		\includegraphics[width=2.7cm,height=1.58cm]{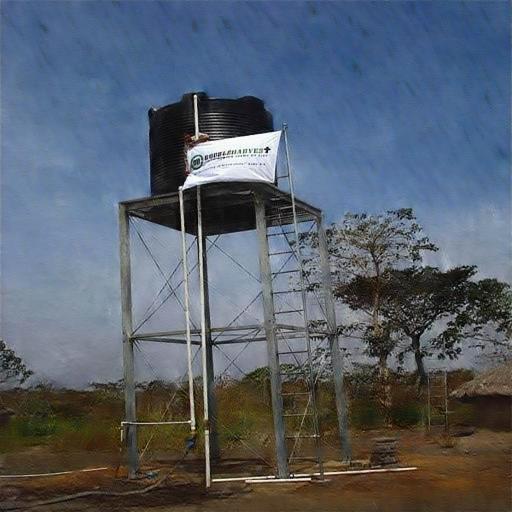}
		\captionsetup{labelformat=empty}
		\captionsetup{justification=centering}
	\end{minipage}
	\begin{minipage}{.16\textwidth}
		\centering
		\caption*{\emph{ PSNR: \textbf{28.29} \\SSIM: \textbf{0.93}}}
		\vskip-10pt
		\includegraphics[width=2.7cm,height=1.58cm]{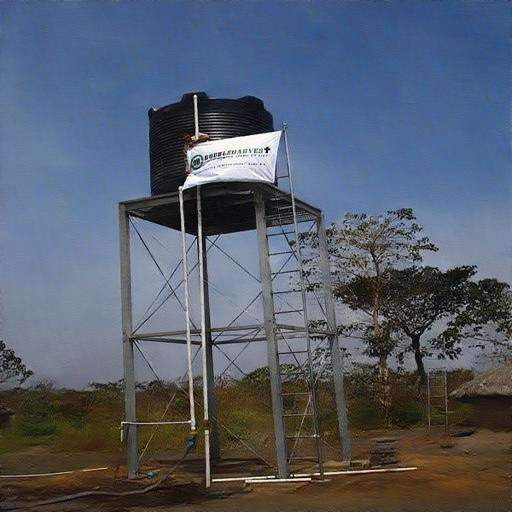}
		\captionsetup{labelformat=empty}
		\captionsetup{justification=centering}
	\end{minipage} 
	\begin{minipage}{.16\textwidth}
		\centering
		\caption*{\emph{PSNR: Inf\\SSIM: 1}}
		\vskip-10pt
		\includegraphics[width=2.7cm,height=1.58cm]{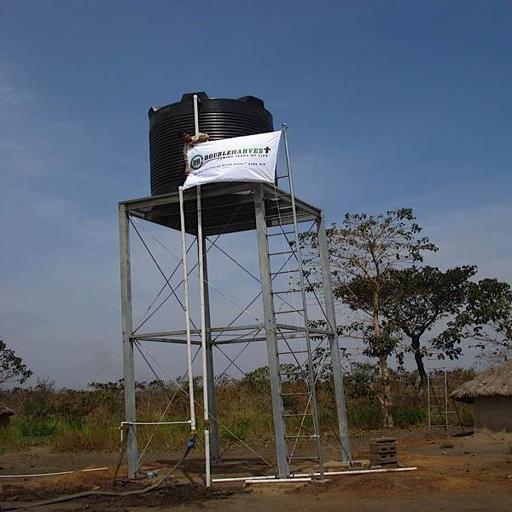}
		\captionsetup{labelformat=empty}
		\captionsetup{justification=centering}
	\end{minipage}\\        \vskip+6pt
	\begin{minipage}{.16\textwidth}
		\centering
		\caption*{\emph{PSNR:14.35 \\SSIM: 0.63}}
		\vskip-10pt
		\includegraphics[width=2.7cm,height=1.58cm]{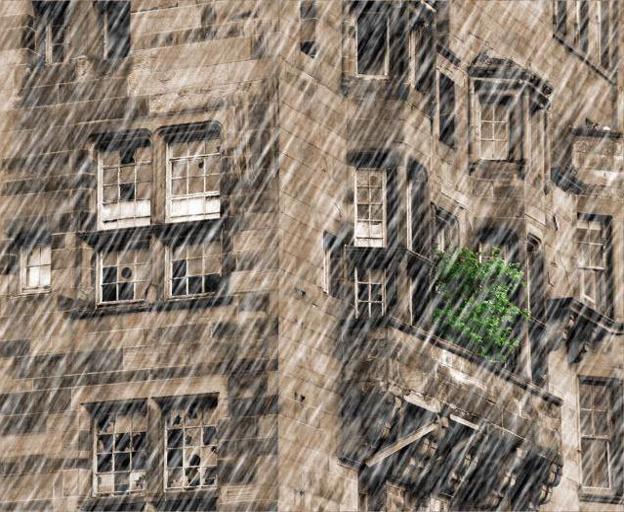}
		\captionsetup{labelformat=empty}
		\captionsetup{justification=centering}
	\end{minipage} 
	\begin{minipage}{.16\textwidth}
		\centering
		\caption*{\emph{PSNR:15.75 \\SSIM: 0.71}}
		\vskip-10pt
		\includegraphics[width=2.7cm,height=1.58cm]{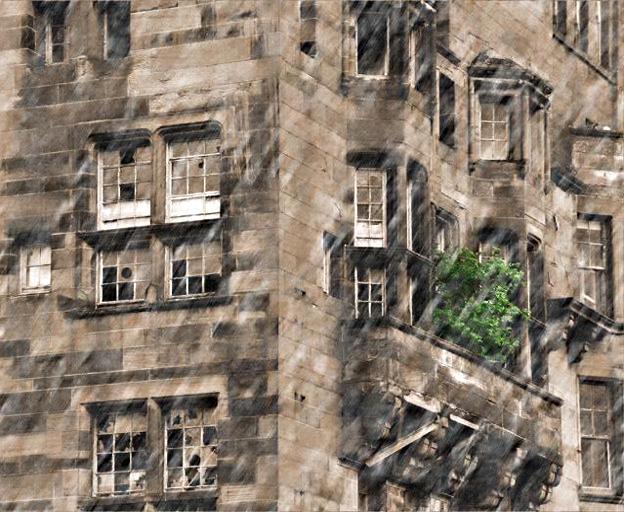}
		\captionsetup{labelformat=empty}
		\captionsetup{justification=centering}
		
	\end{minipage} 
	\begin{minipage}{.16\textwidth}
		\centering
		\caption*{\emph{PSNR:23.01\\SSIM:0.79}}
		\vskip-10pt
		\includegraphics[width=2.7cm,height=1.58cm]{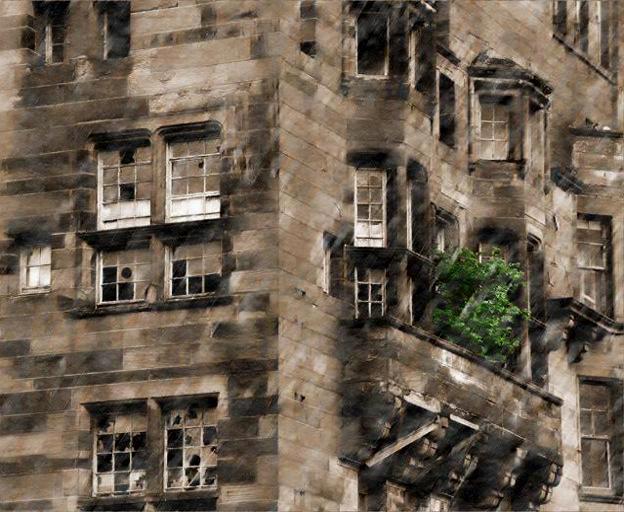}
		\captionsetup{labelformat=empty}
		\captionsetup{justification=centering}
	\end{minipage} 
	\begin{minipage}{.16\textwidth}
		\centering
		\caption*{\emph{PSNR: 27.52 \\SSIM:0.90}}
		\vskip-10pt
		\includegraphics[width=2.7cm,height=1.58cm]{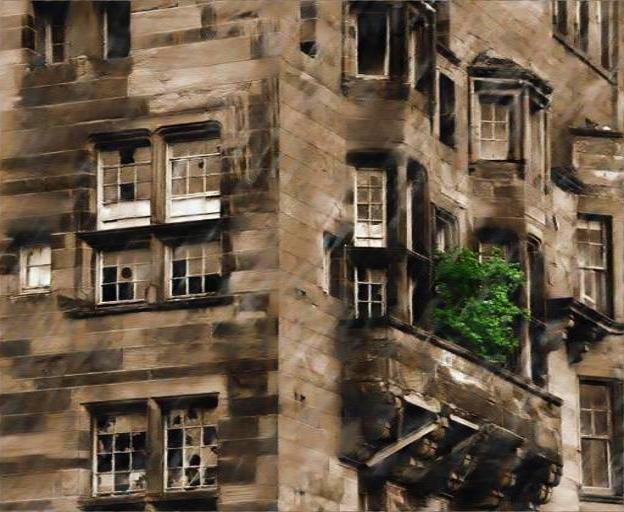}
		\captionsetup{labelformat=empty}
		\captionsetup{justification=centering}
	\end{minipage} 
	\begin{minipage}{.16\textwidth}
		\centering
		\caption*{\emph{PSNR: \textbf{28.83}\\SSIM:\textbf{0.92}}}
		\vskip-10pt
		\includegraphics[width=2.7cm,height=1.58cm]{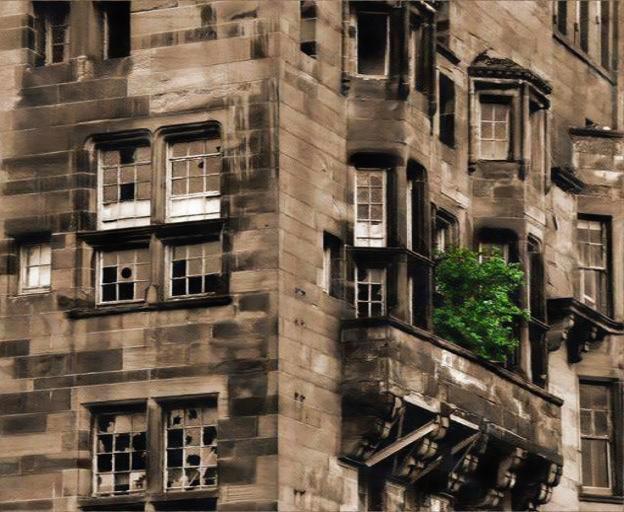}
		\captionsetup{labelformat=empty}
		\captionsetup{justification=centering}
	\end{minipage}
	\begin{minipage}{.16\textwidth}
		\centering
		\caption*{\emph{PSNR: Inf \\SSIM:1}}
		\vskip-10pt
		\includegraphics[width=2.7cm,height=1.58cm]{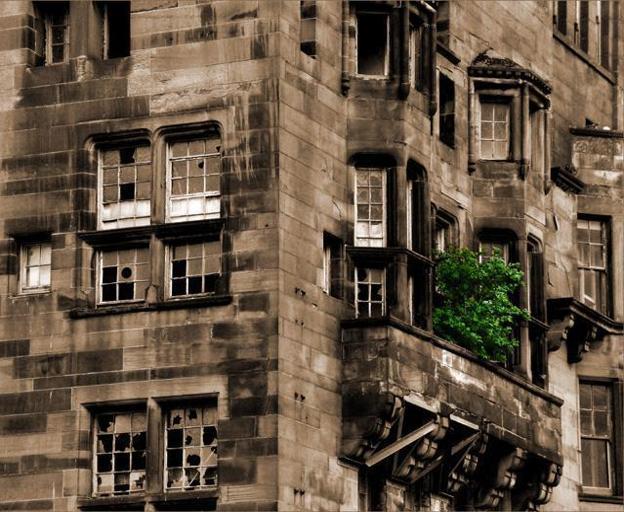}
		\captionsetup{labelformat=empty}
		\captionsetup{justification=centering}
	\end{minipage} \\    \vskip+6pt
	\begin{minipage}{.16\textwidth}
		\centering
		\caption*{\emph{PSNR:20.57 \\SSIM: 0.83}}
		\vskip-10pt
		\includegraphics[width=2.7cm,height=1.58cm]{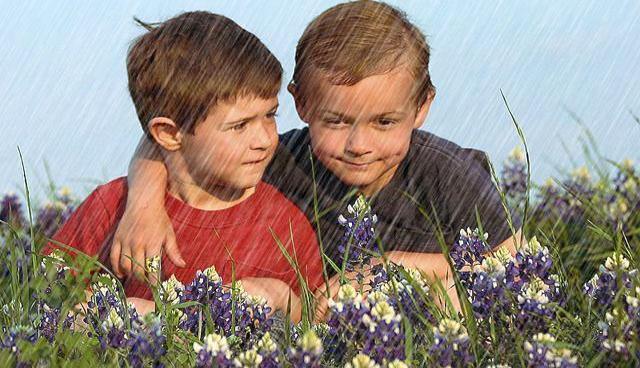}
		\captionsetup{labelformat=empty}
		\captionsetup{justification=centering}
		\caption*{\emph{Rainy Image} \\ \quad\\ \quad}
	\end{minipage} 
	\begin{minipage}{.16\textwidth}
		\centering
		\caption*{\emph{PSNR:22.23 \\SSIM: 0.90}}
		\vskip-10pt
		\includegraphics[width=2.7cm,height=1.58cm]{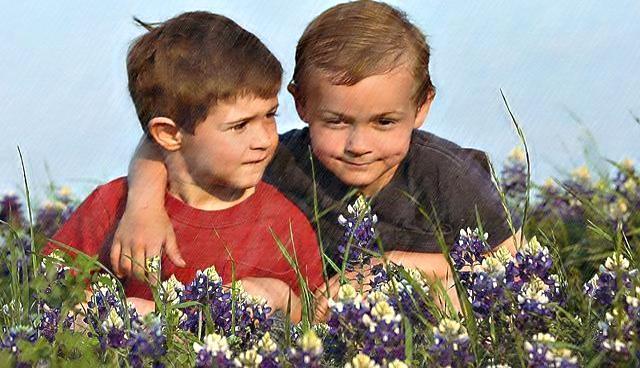}
		\captionsetup{labelformat=empty}
		\captionsetup{justification=centering}
		\caption*{\emph{Fu et al. \cite{Authors17d}(TIP'17)} \\ \quad}
	\end{minipage} 
	\begin{minipage}{.16\textwidth}
		\centering
		\caption*{\emph{PSNR:26.41\\SSIM:0.93}}
		\vskip-10pt
		\includegraphics[width=2.7cm,height=1.58cm]{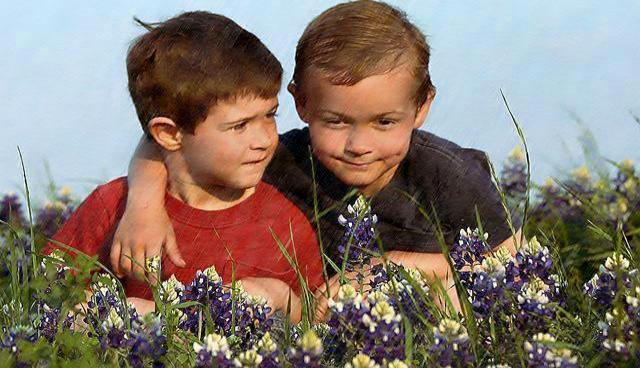}
		\captionsetup{labelformat=empty}
		\captionsetup{justification=centering}
		\caption*{\emph{DDN \cite{Authors17f}(CVPR'17)} \\ \quad}
	\end{minipage} 
	\begin{minipage}{.16\textwidth}
		\centering
		\caption*{\emph{PSNR: 27.23 \\SSIM:0.95}}
		\vskip-10pt
		\includegraphics[width=2.7cm,height=1.58cm]{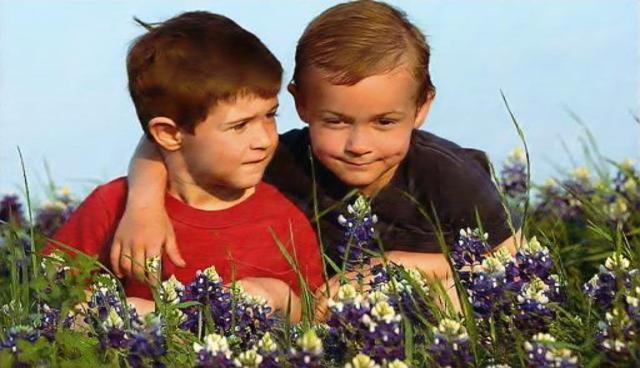}
		\captionsetup{labelformat=empty}
		\captionsetup{justification=centering}
		\caption*{\emph{DID-MDN \cite{Authors18}(CVPR'18)} \\ \quad}
	\end{minipage} 
	\begin{minipage}{.16\textwidth}
		\centering
		\caption*{\emph{PSNR: \textbf{28.21}\\SSIM:\textbf{0.95}}}
		\vskip-10pt
		\includegraphics[width=2.7cm,height=1.58cm]{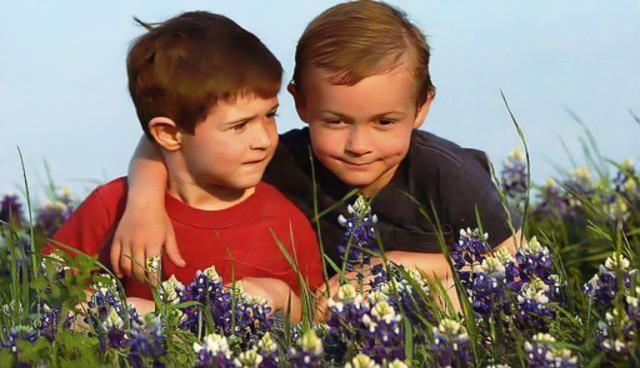}
		\captionsetup{labelformat=empty}
		\captionsetup{justification=centering}
		\caption*{\emph{Ours} \\ \quad\\ \quad}
	\end{minipage}
	\begin{minipage}{.16\textwidth}
		\centering
		\caption*{\emph{PSNR: Inf \\SSIM:1}}
		\vskip-10pt
		\includegraphics[width=2.7cm,height=1.58cm]{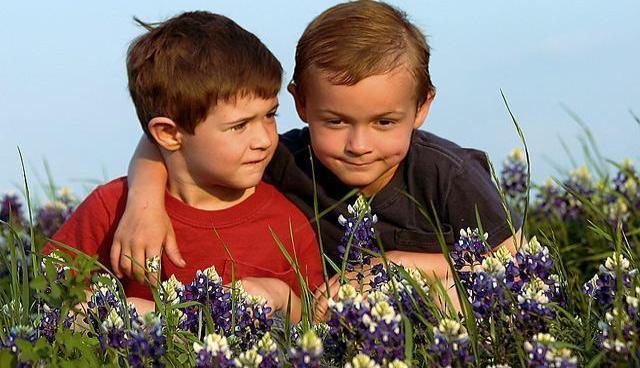}
		\captionsetup{labelformat=empty}
		\captionsetup{justification=centering}
		\caption*{\emph{Ground Truth} \\ \quad\\ \quad}
	\end{minipage} \\
	\vskip -30pt
	\caption{De-rained results on synthetic datasets \textit{Test-1} and \textit{Test-2} consisting  different rain levels (low, medium and heavy) and different directions.}\label{Fig:exp4}
	\vskip-10pt
\end{figure*}

We preformed similar experiments to see how much improvement cycle spinning brings over DDN \cite{Authors17f} and DID-MDN \cite{Authors18}.  In general, we observe approximately $0.25$ dB gain in the performance with cycle spinning compared to without cycle spinning as shown in Table \ref{cyclesp}.

\begin{table}[htp!]
	\vskip-10pt
	\caption{PSNR and SSIM (PSNR$|$SSIM) results corresponding to the ablation study regarding the use of cycle spinning.}
	\vskip-10pt
	\resizebox{8.25cm}{!}{
		\label{cyclesp}
		\begin{tabular}{|c|c|c|c|c|c|c|c|}
			\hline
			Dataset & \begin{tabular}[c]{@{}c@{}}Rainy\\ Image\end{tabular} &  DDN \cite{Authors17f} & \begin{tabular}[c]{@{}c@{}}DDN \cite{Authors17f} + \\ cycle spinning\end{tabular} & DID-MDN \cite{Authors18} & \begin{tabular}[c]{@{}c@{}}DID-MDN \cite{Authors18} + \\ cycle spinning\end{tabular} & UMRL & \begin{tabular}[c]{@{}c@{}}UMRL+\\ cycle spinning\end{tabular} \\ \hline
			\textit{Test-1} & 21.15$|$0.77 & 27.33$|$0.90 & 27.52$|$0.91  & 27.95$|$0.91 & 28.19$|$0.91 & 29.42$|$0.91 & \textbf{29.77$|$0.92} \\ \hline
			\textit{Test-2} & 19.31$|$0.77 & 25.63$|$0.88 & 25.90$|$0.89 & 26.08$|$0.90  &  26.37$|$0.91 & 26.47$|$0.91 & \textbf{26.67$|$0.92} \\ \hline
		\end{tabular}
	}
	\vskip-10pt
\end{table}

Figure \ref{Fig:exp3} illustrates that confidence map is guiding the network to learn the rain content at the edges and texture regions clearly by imposing low confidence values. From Figure \ref{Fig:exp3} by looking at the histograms of confidence maps at different scales, we can observe that as the scale is increasing the confidence values are approaching 1 at most of the pixels.  This behavior is expected since at lower scales, the rain streaks will be blurry (see Figure \ref{Fig:residual}) and the network is less confident about the values it estimates. This explains why UMRL tries to increase the confidence value by estimating accurate residual maps, in return CN is computing and feeding back the possible areas where UMRL goes wrong.

\begin{table*}[ht!]
	\begin{center}
		\caption{\large{PSNR and SSIM comparison of UMRL against state-of-art methods (PSNR$|$SSIM))}}
		\vskip-10pt
		\resizebox{\textwidth}{!}{
			\label{syn_comp}
			\begin{tabular}{|c|c|c|c|c|c|c|c|c|}
				
				\hline
				Dataset & \begin{tabular}[c]{@{}c@{}}Rainy\\ Image\end{tabular} & \begin{tabular}[c]{@{}c@{}}GMM based\\ \cite{Authors16}(CVPR'16)\end{tabular} & \begin{tabular}[c]{@{}c@{}}Fu et al.\\ \cite{Authors17d}(TIP'17)\end{tabular} & \begin{tabular}[c]{@{}c@{}}JORDER\\ \cite{Authors17h}(CVPR'17)\end{tabular} & \begin{tabular}[c]{@{}c@{}}DDN\\ \cite{Authors17f}(CVPR'17)\end{tabular} & \begin{tabular}[c]{@{}c@{}}JBO\\ \cite{Authors17c}(ICCV'17)\end{tabular} &  \begin{tabular}[c]{@{}c@{}}DID-MDN\\ \cite{Authors18}(CVPR'18)\end{tabular} & \begin{tabular}[c]{@{}c@{}}UMRL+\\ cycle spinning\end{tabular} \\ \hline
				\textit{Test-1} & 21.15$|$0.77 & 22.75$|$0.84 & 22.07$|$0.84   & 24.32$|$0.86 & 27.33$|$0.90 & 23.05$|$0.85 & 27.95$|$0.91 & \textbf{29.77$|$0.92} \\ \hline
				\textit{Test-2} & 19.31$|$0.77 & 22.60$|$0.81 & 19.73$|$0.83  &  22.26$|$0.84 & 25.63$|$0.88 & 22.45$|$0.84 & 26.08$|$0.90 & \textbf{26.67$|$0.92} \\ \hline
			\end{tabular}
		}
	\end{center}
	\vskip-25pt
\end{table*}

\subsection{Results on Synthethic Test Images}

\begin{figure*}[ht!]
	\begin{center}
		\includegraphics[width=0.195\textwidth,height=0.12\textwidth]{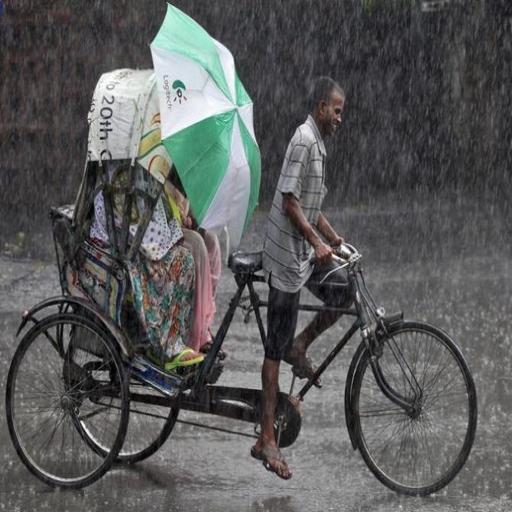}
		\includegraphics[width=0.195\textwidth,height=0.12\textwidth]{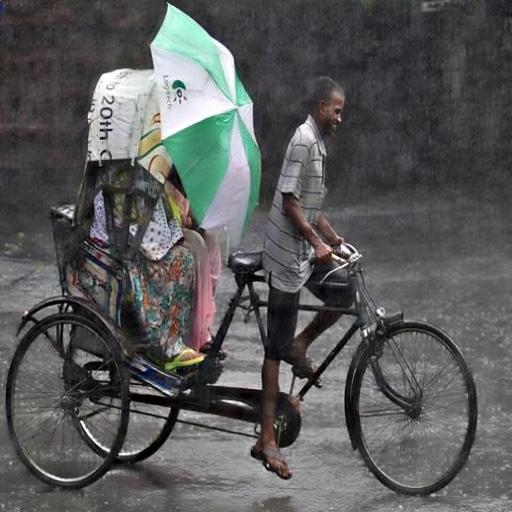}
		\includegraphics[width=0.195\textwidth,height=0.12\textwidth]{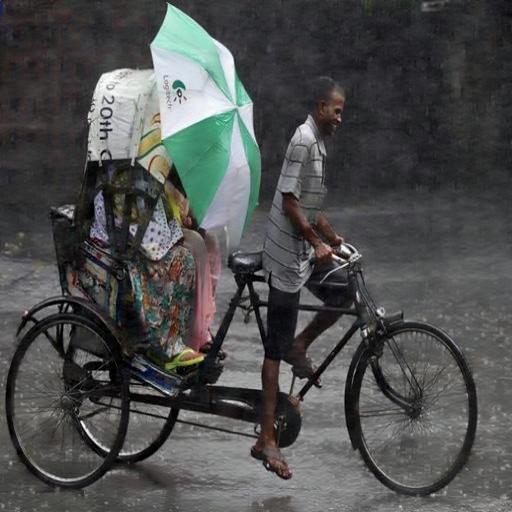}
		\includegraphics[width=0.195\textwidth,height=0.12\textwidth]{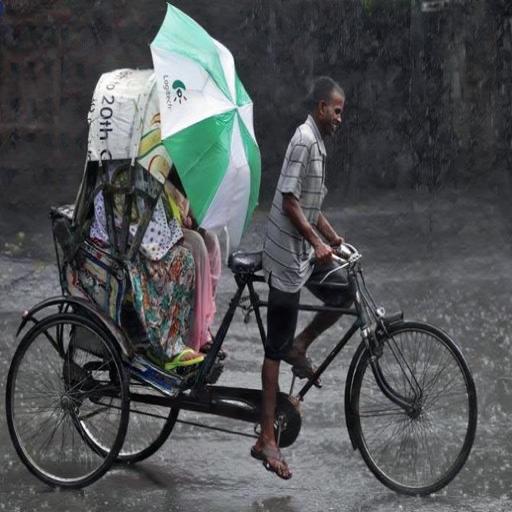}
		\includegraphics[width=0.195\textwidth,height=0.12\textwidth]{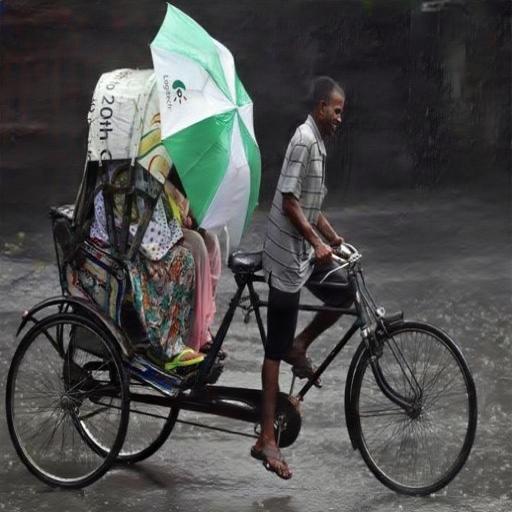}\\
		\includegraphics[width=0.195\textwidth,height=0.12\textwidth]{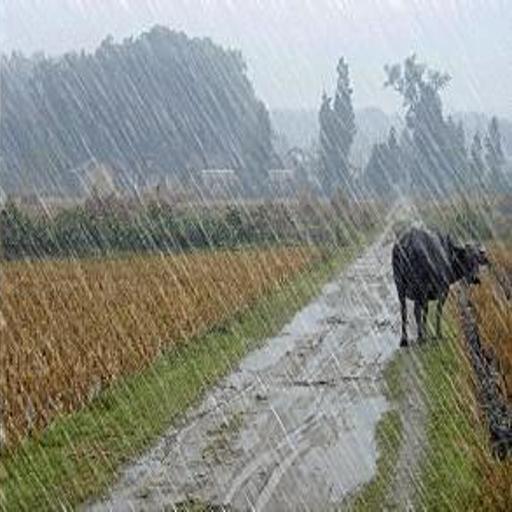}
		\includegraphics[width=0.195\textwidth,height=0.12\textwidth]{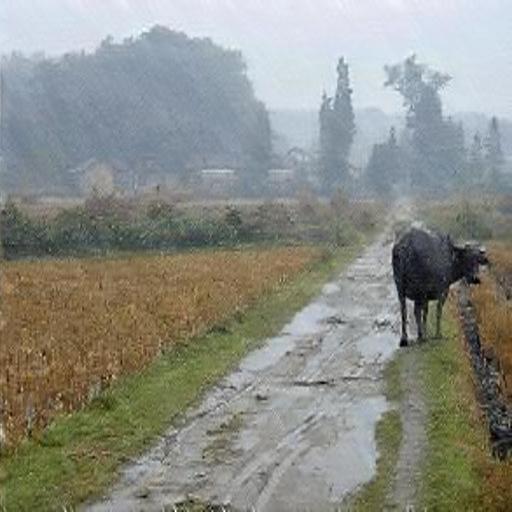}
		\includegraphics[width=0.195\textwidth,height=0.12\textwidth]{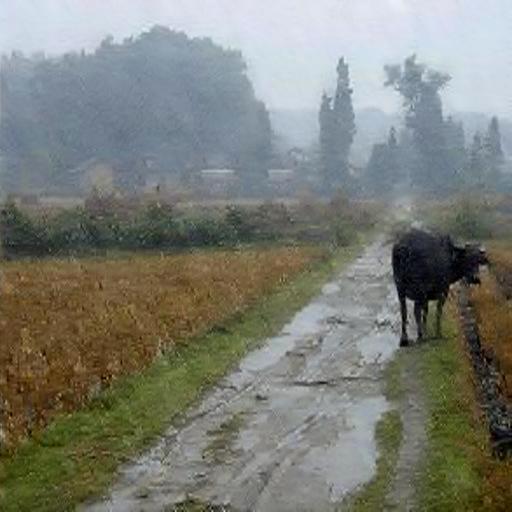}
		\includegraphics[width=0.195\textwidth,height=0.12\textwidth]{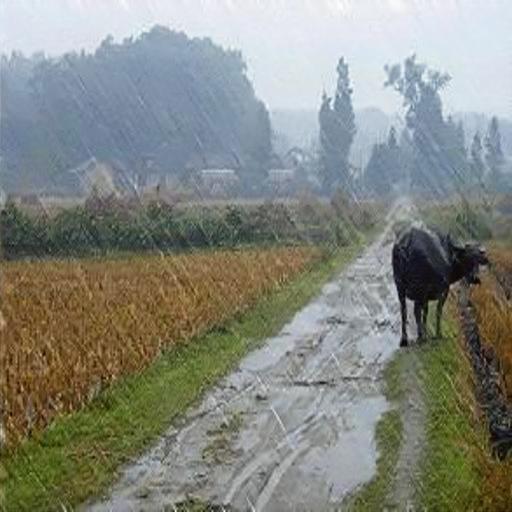}
		\includegraphics[width=0.195\textwidth,height=0.12\textwidth]{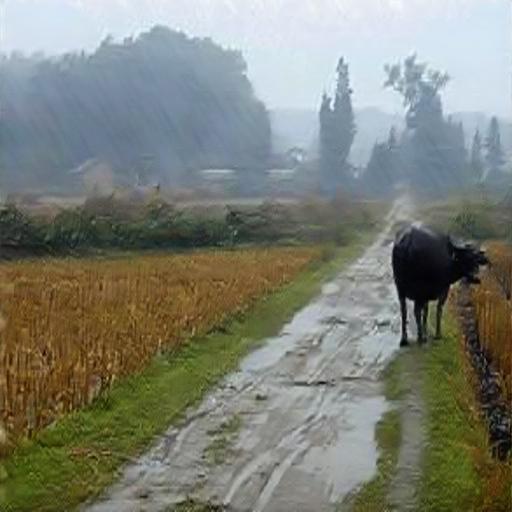}\\
		\includegraphics[width=0.095\textwidth,height=0.06\textwidth]{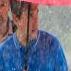}
		\includegraphics[width=0.095\textwidth,height=0.06\textwidth]{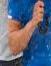}
		\includegraphics[width=0.095\textwidth,height=0.06\textwidth]{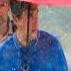}
		\includegraphics[width=0.095\textwidth,height=0.06\textwidth]{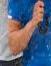}
		\includegraphics[width=0.095\textwidth,height=0.06\textwidth]{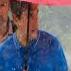}
		\includegraphics[width=0.095\textwidth,height=0.06\textwidth]{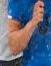}
		\includegraphics[width=0.095\textwidth,height=0.06\textwidth]{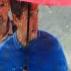}
		\includegraphics[width=0.095\textwidth,height=0.06\textwidth]{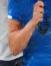}
		\includegraphics[width=0.095\textwidth,height=0.06\textwidth]{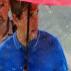}
		\includegraphics[width=0.095\textwidth,height=0.06\textwidth]{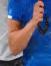}\\
		\includegraphics[width=0.195\textwidth,height=0.12\textwidth]{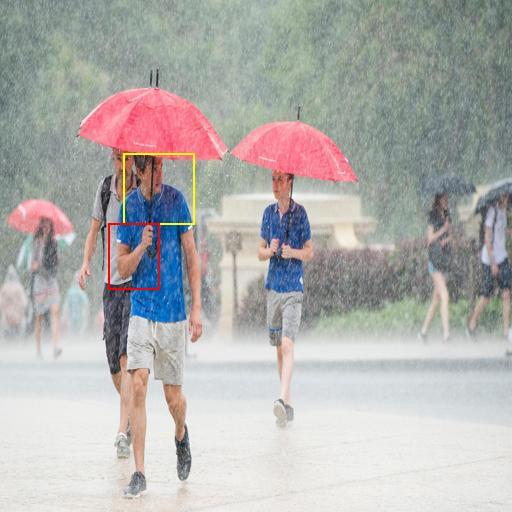}
		\includegraphics[width=0.195\textwidth,height=0.12\textwidth]{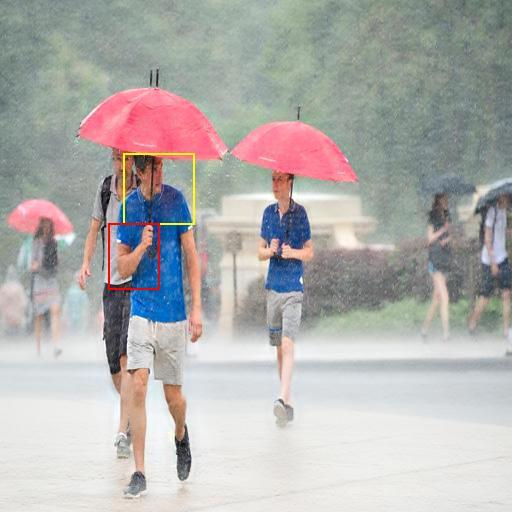}
		\includegraphics[width=0.195\textwidth,height=0.12\textwidth]{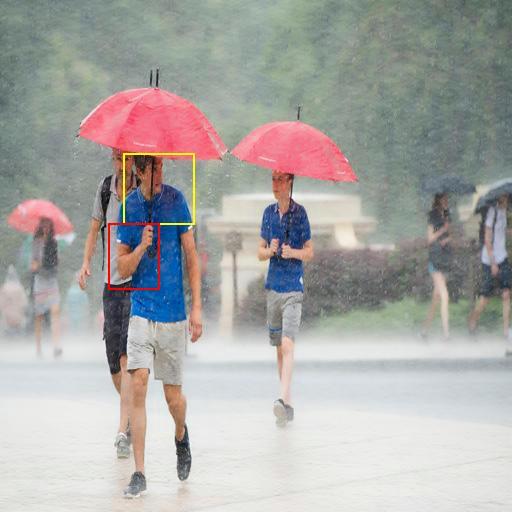}
		\includegraphics[width=0.195\textwidth,height=0.12\textwidth]{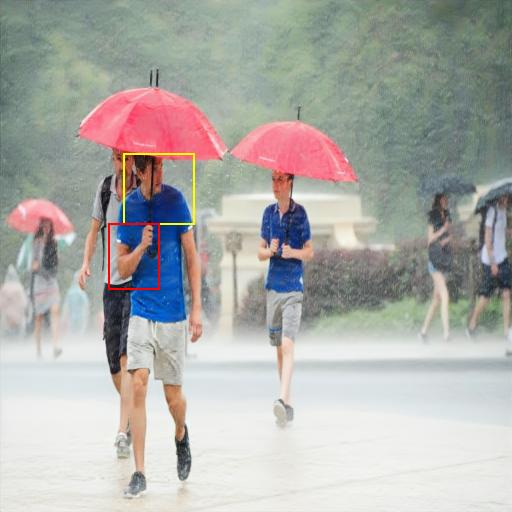}
		\includegraphics[width=0.195\textwidth,height=0.12\textwidth]{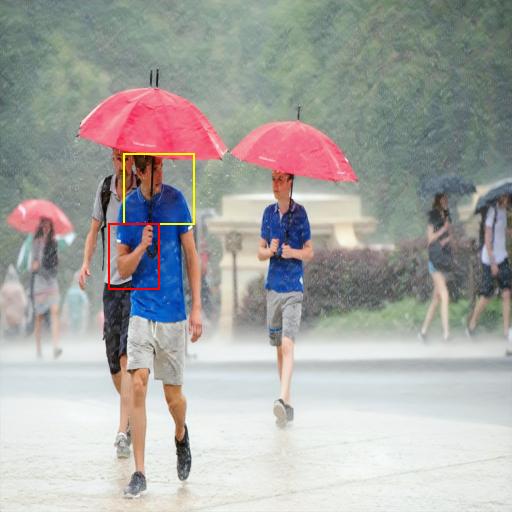}\\
		\textit{ Rainy Image \hskip50pt Fu et al. \hskip50pt DDN \hskip80pt DID-MDN  \hskip50pt Ours\\}
		\textit{\cite{Authors17d}(TIP'17) \hskip40pt \cite{Authors17f}(CVPR'17) \hskip40pt \cite{Authors18}(CVPR'18)\\}
		\caption{De-rained results on sample real-world images.}
		\label{Fig:exp6}
	\end{center}
	\vskip-25pt
\end{figure*}

The proposed UMRL method based on cycle spinning is compared against the state-of-the-art algorithms qualitatively and quantitatively. Table \ref{syn_comp} shows the quantitative performance of our method.  As it can be seen from this table, our method clearly out-performs the present state-of-the-art algorithms.  Furthermore, we compare our method against a recent ECCV'18 method called REcurrent SE Context Aggregation Net (RESCAN) \cite{Authors18d} using the \textit{Rain800} dataset containing 100 images from \cite{Authors17e}.  The PSNR and SSIM values achieved by RESCAN \cite{Authors18d} are 24.37 and 0.84, whereas our method achieved 24.59 and 0.87, respectively.

Figure \ref{Fig:exp4}  shows the qualitative performance of different methods on three sample images from \textit{Test-1} and \textit{Test-2} datasets. Though Fu et al. (TIP'17) \cite{Authors17d} is able to remove some rain streaks, it is unable to remove all the rain components. DDN \cite{Authors17f} is over de-raining on some images and on others it is slightly under de-raining as shown in the third column of Figure \ref{Fig:exp4}. DID-MDN \cite{Authors18} is over de-raining as shown in the fourth column of Figure \ref{Fig:exp4} where it removes the texture on wooden wall, edges of the building in second image.  Furthermore, it blurs the edges of water tank in the fourth image. By comparing third and fourth images of the fourth column, we see that the outputs of DID-MDN \cite{Authors18} has a small blurred version of the residual streaks in the sky of those images. Visually we can see in the fifth column of Figure \ref{Fig:exp4}, our method produces images without any artifacts. For example in (i) it is able to recover the texture on wooden wall,  in (ii) it is able to produce images with clear sky in the third and fourth images of fifth column, and in (iii) it is able to produce the sharp edges in second and fourth images.

To de-rain an image of size $512\times 512$, on average UMRL takes about 0.05 seconds, and UMRL with cycle spinning takes about 5.1 seconds.

\subsection{Results on Real-World Rainy Images}

We conducted experiments on the real-world images provided by \cite{Authors17e,Authors17d,Authors18}.  Results are shown in  Figure \ref{Fig:exp6}. Similar to the results obtained on synthetic images, we observe the same trend of either over de-raining or under de-raining by the other methods. On the other hand, our method is able to remove rain streaks while preserving details of objects in the resultant output images.  For example, the background and man's face in the first image of the fifth column is more clear than the outputs from other methods. Also, Trees and plants in the second image of the fifth column, front man's face and t-shirt collar in the third image are visually more clear than the results from other method. All of these experiments clearly show that our method can handle different levels of rain (low, medium and high) with different shapes and scales. More results on synthetic and real-world images are provided in the supplementary material.

\section{Conclusion}
We proposed a novel UMRL method based on cycle spinning to address the single image de-raining problem. In our approach, we introduced uncertainty guided residual learning where the network tries to learn the residual maps and the corresponding confidence maps at different scales which were then fed back to the subsequent layers to guide the network. In addition to UMRL, we analyzed the benefits of using  cycle spinning in de-raining using various recently proposed deep de-raining networks. Extensive experiments showed that UMRL is robust enough to handle different levels of rain content for both synthetic and real-world rainy images. 

\section*{Acknowledgements}
This research is based upon work supported by the Office of the Director of National Intelligence (ODNI), Intelligence Advanced Research Projects Activity (IARPA), via IARPA R$\&$D Contract No. 2014-14071600012. The views and conclusions contained herein are those of the authors and should not be interpreted as necessarily representing the official policies or endorsements, either expressed or implied, of the ODNI, IARPA, or the U.S. Government.
{\small
	\bibliographystyle{ieee}
	\bibliography{Derain_CVPR19}
}
	
\end{document}